\documentclass[journal, 10pt]{IEEEtran}
\usepackage{amsmath,amsfonts}
\usepackage{array}
\usepackage{textcomp}
\usepackage{stfloats}
\usepackage{url}
\usepackage{verbatim}
\usepackage{graphicx}
\usepackage{enumitem}
\usepackage{url}
\usepackage{graphics}
\usepackage{tikz}
\usepackage{multirow}
\usepackage{graphicx}
\usepackage{tabularx}
\usepackage{listings}
\usepackage{enumitem}
\usepackage{makecell}
\usepackage{algpseudocode}
\usepackage{amsmath}
\usepackage{amssymb}
\usepackage[all]{nowidow}

\usepackage{mathtools}
\usepackage{multirow}
\usepackage{algpseudocode}
\usepackage{url}

\usepackage[utf8]{inputenc}
\usepackage{enumitem}
\usepackage{url}
\usepackage{tikz}
\usepackage{caption}
\usepackage{subcaption}
\usepackage{multirow}
\usepackage{enumitem}
\usepackage{tabularx}
\usepackage{multirow}
\usepackage{amsthm}
\usepackage{listings}
\usepackage{float}
\usepackage{comment}
\usepackage{hyperref}
\usepackage{amsmath}
\usepackage[ruled,lined,boxed,linesnumbered]{algorithm2e}
\usepackage{amsthm}
\usepackage{balance}

\usepackage{titlesec}

\usepackage{caption}

\newcounter{reviewercount}
\newcounter{commentcount}

\newcommand{\aeditor}%
  {\bigskip\noindent {\bf COMMENTS OF THE ASSOCIATE EDITOR}%
  \setcounter{commentcount}{0}\par 
}

\usepackage[utf8]{inputenc}
\usepackage{enumitem}
\usepackage{url}
\usepackage{multirow}
\usepackage{graphicx}
\usepackage{enumitem}
\usepackage{tabularx}
\usepackage{multirow}
\usepackage{amsthm}
\usepackage{listings}
\usepackage{float}
\usepackage{comment}
\usepackage[ruled,lined,boxed,linesnumbered]{algorithm2e}

\usepackage{soul}

\usepackage{amsthm}
\theoremstyle{plain}

\theoremstyle{definition}

\newtheoremstyle{claim}
  {\topsep}
  {\topsep}
  {}
  {}
  {\itshape}
  {}
  {.5em}
  {\thmname{#1}\thmnumber{ #2}\thmnote{ (#3)}}

\SetKwInput{KwOutput}{Output}
\SetKwInput{KwInput}{Input} 

\graphicspath{ {FIGS/} }

\begin{document}

\setcounter{page}{1}
\twocolumn

\title{K-SpecPart: Supervised embedding algorithms and cut overlay for improved hypergraph partitioning}
\author{Ismail Bustany,~\IEEEmembership{Member,~IEEE,} 
        Andrew~B.~Kahng,~\IEEEmembership{Fellow,~IEEE,}
        Ioannis Koutis, 
        \\Bodhisatta Pramanik,~\IEEEmembership{Student Member,~IEEE}
        and~Zhiang~Wang,~\IEEEmembership{Student Member,~IEEE}}

\maketitle

\begin{abstract}
State-of-the-art hypergraph partitioners follow the multilevel paradigm 
that constructs multiple levels of progressively 
coarser hypergraphs that are used to drive \textcolor{black}{cut} refinement
on each level of the hierarchy. 
Multilevel partitioners are subject to two limitations: 
(i) hypergraph coarsening processes rely on local neighborhood structure 
without fully considering the global structure of the hypergraph;
and (ii) refinement heuristics risk entrapment in local minima.
\textcolor{black}{In this paper, we describe {\em K-SpecPart}, 
a supervised spectral framework for multi-way partitioning that directly tackles these two limitations. \textit{K-SpecPart} relies on the computation of generalized eigenvectors and supervised dimensionality reduction techniques to generate vertex embeddings. These are computational primitives that are not only fast, but \textcolor{black}{embeddings} also capture global structural properties of the hypergraph that are not explicitly considered by existing partitioners.} \textit{K-SpecPart} then converts the vertex \textcolor{black}{embeddings} into multiple partitioning solutions. Unlike multilevel partitioners that only \textcolor{black}{consider} the best solution, \textit{K-SpecPart} introduces the idea of ``ensembling'' \textcolor{black}{multiple} solutions via a {\em cut-overlay clustering} technique that often enables the use of computationally demanding partitioning methods such as ILP (integer linear programming). Using the output of a standard partitioner as a supervision hint, \textit{K-SpecPart} effectively combines the strengths of established multilevel partitioning techniques with the benefits of spectral graph theory and other combinatorial algorithms. \textcolor{black}{\textit{K-SpecPart} significantly extends ideas and algorithms that first appeared in our previous work on the bipartitioner \textit{SpecPart} [Bustany~et~al.~ICCAD 2022]. Our experiments demonstrate the effectiveness of {\em K-SpecPart}.
For bipartitioning, {\em K-SpecPart} produces solutions
with up to $\sim$15\% cutsize improvement over {\em SpecPart}. For multi-way partitioning,  {\em K-SpecPart} produces solutions with up to $\sim$20\% cutsize improvement over leading partitioners {\em hMETIS} and {\em KaHyPar}.}
\end{abstract}

\section{Introduction}
\label{sec:intro}

Balanced hypergraph partitioning is a well-studied, fundamental combinatorial optimization problem with multiple applications in EDA. The objective is to partition vertices of a hypergraph into a specified number of disjoint {\em blocks} such that each block has bounded size and the {\em cutsize}, i.e., the number of hyperedges spanning multiple blocks, is minimized~\cite{SebastianTLYCP22}. 

Many hypergraph partitioners have been proposed over the past decades. State-of-the-art partitioners, including {\em MLPart}~\cite{CaldwellKM00}, {\em PaToH}~\cite{CatalyürekA11}, {\em KaHyPar}~\cite{SebastianTLYCP22} and {\em hMETIS}~\cite{KarypisAKS99},
follow the multilevel paradigm~\cite{KarypisAKS99}. Another thread of work that has been less successful in practice uses variants of unsupervised spectral clustering~\cite{ZienSC99, HagenK91, RebagliatiV11, AlpertK95}. 
All partitioning algorithms that are constrained by practical runtime constraints are inevitably bound to limitations, due to the computational complexity of the problem. However, different types of algorithms may have complementary strengths. For example, multilevel algorithms attempt to directly optimize the combinatorial objective, but they are bound by the local nature of their clustering heuristics and the entrapment in local minima that cannot be circumvented by their greedy refinement heuristics~\cite{FiducciaM82, HeuerSS19}. On the other hand, spectral algorithms by design take into account global properties of the hypergraph, albeit at the expense of optimizing surrogate objectives that may introduce significant approximation error.

\textit{K-SpecPart} is based on a novel general concept: a partitioning solution is viewed as a {\em hint} that can be used as input to supervised algorithms. The idea enables us to combine the strengths of established partitioning techniques with the benefits of supervised methods, and in particular spectral algorithms. \textcolor{black}{Following} are our main algorithmic and experimental contributions.

\noindent $\bullet$ \textcolor{black}{\textit{Supervised Spectral $K$-way Embedding.} \\ Similar to \textit{SpecPart}~\cite{SpecPart}, \textit{K-SpecPart} adapts the supervised spectral algorithm of~\cite{CucuringuKCMP16} to generate a vertex embedding by solving a generalized eigenvalue problem. Spectral $K$-way partitioning usually involves either the computation of $K$ eigenvectors of a single problem, or recursive bipartitioning. In our work, the availability of the $K$-way hint leads to a ``one-vs-rest'' approach that involves three fundamental steps. (i) We extract multiple two-way partitioning solutions from a K-way {\em hint} partitioning solution, and incorporate these as hint solutions into multiple instances of the generalized eigenvalue problem. (ii) We subsequently solve the problem instances to generate multiple eigenvectors. (iii ) The (column) eigenvectors from these instances are horizontally stacked to form a large-dimensional embedding. This particular way of generating a supervised $K$-way embedding is novel and may be of independent interest. [Section \ref{sec:embedding}]}

\noindent $\bullet$ \textcolor{black}{\textit{Supervised Dimensionality Reduction.} \\
\textit{K-SpecPart} generates embeddings that have larger dimensions than those in \textit{SpecPart}, posing a computational bottleneck for subsequent steps. To mitigate this problem, we use {\em linear discriminant analysis (LDA)}, a {\em supervised} dimensionality reduction technique, where we leverage again the $K$-way hint. This produces a low-dimensional embedding that respects (spatially) the $K$-way hint solution. Our experimental results show that this step not only reduces the  runtime significantly ($\sim$10X), but also slightly improves ($\sim$1\%) the cutsize, relative to using the large-dimensional embedding.
[Sections \ref{sec:embedding} and \ref{sec:lda}]}
 
\noindent $\bullet$  \textit{Cut Distilling Trees and Tree Partitioning.} \\
Converting a vertex embedding to a $K$-way partitioning is an integral step of \textit{K-SpecPart}. \textit{SpecPart} introduced in this context a novel approach that uses the hypergraph and the vertex embedding to compute a family of weighted trees that in some sense \textit{distill} the cut structure of the hypergraph. This effectively reduces the hypergraph partitioning problem to a $K$-way tree partitioning problem. \textcolor{black}{Of course, \textit{$K$ $>$ $2$} makes for a significantly more challenging problem, \textcolor{black}{which} we tackle in \textit{K-SpecPart}. More specifically, we use recursive bipartitioning by extending the tree partitioning algorithm of \cite{SpecPart} and augmenting it with a refinement step using the multi-way Fiduccia-Mattheyses (FM) algorithm~\cite{FiducciaM82, HeuerSS19}. This step is essentially an encapsulated use of an established partitioning algorithm, tapping again into the power of \textcolor{black}{existing} methods. [Section \ref{sec:tree_construction}]}

\noindent $\bullet$  \textit{Cut Overlay and Optimization.} \\ \textit{K-SpecPart} is an iterative algorithm that uses its partitioning solution from iteration $i$ as a hint for subsequent iteration $i+1$. Standard multilevel partitioners \textcolor{black}{compute} multiple solutions and pick the best while discarding the rest. \textit{K-SpecPart}, however, uses its entire pool of computed solutions in order to find a further improved partitioning solution, via a solution ensembling technique, \textit{cut-overlay clustering}~\cite{SpecPart}. Specifically, we extract clusters by removing from the hypergraph the union of the hyperedges cut by any partitioning solution in the pool. The resulting clustered hypergraph typically comprises only hundreds of vertices, enabling ILP-based (integer linear program) hypergraph partitioning to efficiently identify the optimal partitioning of the set of clusters. \textcolor{black}{The solution is then subsequently ``lifted'' to the original hypergraph and further refined with FM.} [Section \ref{sec:overlay_part}]

\noindent $\bullet$  \textit{Autotuning.} \\ We apply autotuning~\cite{Ray} on the hyperparameters of standard partitioners in order to generate a better hint for {\em K-SpecPart}.  Our experiments show that this can further push the leaderboard for \textcolor{black}{well-studied} benchmarks. [Section \ref{sec:autotuning}] 

\noindent $\bullet$  \textcolor{black}{\textit{An Extensive Experimental Study.} \\ We validate {\em K-SpecPart} on multiple benchmark sets  ({\em ISPD98 VLSI Circuit Benchmark Suite}~\cite{Alpert98} and {\em Titan23}~\cite{MurrayWLLB13}) with state-of-the-art partitioners ({\em hMETIS}~\cite{KarypisAKS99} and {\em KaHyPar}~\cite{SebastianTLYCP22}). Experimental results show that for some cases, {\em K-SpecPart} can  improve cutsize by more than $50$\% over {\em hMETIS} and/or {\em KaHyPar} for bipartitioning and by more than $20$\% for multi-way partitioning. [Section \ref{sec:experiments}]. We also conduct a large ablation study in Sections~\ref{sec:ablation} to~\ref{sec:autotuning} that shows how each of the individual components of our algorithm contributes in the overall result. Besides publishing all codes and scripts, we also publish a leaderboard with the best known partitioning solutions for all our benchmark instances in order to motivate future research~\cite{TILOS-HGPart}.} 

\textcolor{black}{{\em K-SpecPart} is built as an extension to {\em SpecPart} but significantly extends the ideas in~\cite{SpecPart}.} This framework includes a variety of novel components that may seem challenging to comply with the strict runtime constraints of practical hypergraph partitioning. However, the choice of numerical solvers~\cite{KoutisMP14,KoutisMT11} along with careful engineering enables a very efficient implementation, with further parallelization potential [Section~\ref{sec:runtime}]. {\em K-SpecPart}'s capacity to include supervision information makes it potentially even more powerful in industrial pipelines. More importantly, its components are subject to individual improvement possibly leveraging machine learning and other optimization-based techniques (Section~\ref{sec:conclusion}). We thus believe that our work may eventually lead to a departure from the  multilevel paradigm that has dominated the field for the past  quarter-century.

\begin{table}[!h]
     \centering
     \resizebox{\columnwidth}{!}{
    \begin{tabular}{|l|l|}
    \hline
    \textbf{Term} & \textbf{Description} \\ \hline
    $H(V, E)$  & Hypergraph $H$ with vertices $V$ and hyperedges $E$ \\ \hline
    \multirow{2}{*}{$H_c(V_c, E_c)$ }  & Clustered hypergraph $H_c$ where each vertex $v_c \in V_c$ \\ 
    & corresponds to a group of vertices in $H(V, E)$ \\ \hline
    $G(V, E)$  & Graph $G$ with vertices $V$ and edges $E$ \\ \hline
    $\tilde{G}$ & Spectral sparsifier of $G$ \\ \hline
    $T(V, E_T)$ & Tree $T$ with vertices $V$ and edges $E_T$ \\ \hline
    $u,v$        & Vertices in $V$ \\ \hline
    $e_{uv}$ & Edge connecting $u$ and $v$ \\ \hline
    $e_T$       & Edge of tree $T$ \\ \hline
    $w_v, w_e$      & Weight of vertex $v$, or hyperedge $e$, respectively \\ \hline
    $K$        & Number of blocks in a partitioning solution \\ \hline
    $S$        & Partitioning solution, $S = \{V_0, V_1, ..., V_{K-1}\}$ \\ 
    \hline
    $\epsilon$ & Allowed imbalance between blocks in $S$ \\ \hline
    $cut(S)$   & Cut of $S$, $cut(S) = \{ e | e \not \subseteq V_i \text{ for any } i \}$ \\ \hline
    $cutsize_H(S)$ & Cutsize of $S$ on (hyper)graph $H$, i.e., sum of $w_e$, $e \in cut(S)$ \\ \hline 
    \textcolor{black}{$X_{emb}$, $X$}   & \textcolor{black}{Vertex embeddings} \\ 
    \hline
    \end{tabular}
    }
    \caption{Notation.}
    \label{tab:term}
\end{table}

\begin{table}[!htb]
     \centering
     \resizebox{\columnwidth}{!}{
    \begin{tabular}{|l|l|}
    \hline
    \textbf{Parameter} & \textbf{Description} (default setting)\\ \hline
    $m$         & Number of eigenvectors ($m = 2$) \\ \hline
    $\delta$    & Number of best solutions ($\delta = 5$) \\ \hline
    $\beta$     &  \textcolor{black}{Number of iterations of {\em K-SpecPart}} ($\beta$ = 2) \\ \hline
    $\zeta$     & Number of random cycles  ($\zeta = 2$)\\ \hline
    $\gamma$     & Threshold of number of hyperedges (\textcolor{black}{$\gamma = 500$}) \\ \hline
    \end{tabular}
    }
    \caption{Parameters of the {\em K-SpecPart} framework.}
    \label{tab:param}
\end{table}

\section{Preliminaries}
\label{sec:prelim}

\subsection{Hypergraph Partitioning Formulation}
\label{sec:problem_statement}

A hypergraph $H(V,E)$ consists of a set of vertices $V$ and a set of hyperedges $E$ where for each $e\in E$, we have $e\subseteq V$. 
We work with weighted hypergraphs, where each vertex $v\in V$ and each hyperedge $e\in E$ 
are associated with positive weights $w_v$ and $w_e$ respectively. 
Given a hypergraph $H$, we define:
\smallskip
\begin{itemize}[itemsep=3pt,topsep=0pt,leftmargin=*]
\item \textit{$K$-way partition:} A collection ${\cal S}=\cup_{i} V_i$ of $K$ vertex blocks $V_i\subseteq V$ 
such that $V_i\cap V_j = \emptyset$ and $\cup_{i=0}^{K-1} V_i = V.$

\item \textit{Vertex set weight:} For $U\subseteq V$, $W_U = \sum_{v\in U} w_v.$

\item \textit{$\epsilon$-balanced $K$-way partition {\cal S}}: A $K$-way partition such that for all $V_i \subseteq {\cal S}$, 
we have $0\leq \frac{1}{K}-\epsilon \leq {W_{V_i}}/{{W_V}}\leq \frac{1}{K}+\epsilon$.

\item $cut_H({\cal S}) = \{e |  e \not \subseteq V_i  \text{ for all } V_i\subseteq {\cal S} \}. $

\item $cutsize_H({\cal S}) = \sum_{ e\in cut_H({\cal S})}w_e.$

\end{itemize}
\smallskip

The hypergraph partitioning problem 
seeks an $\epsilon$-balanced $K$-way partition $\cal S$ that minimizes $cutsize_H({\cal S}).$

\subsection{Laplacians, Cuts and Eigenvectors} 
\label{sec:rq}

Suppose $G=(V,E,w)$ is a weighted graph. The Laplacian 
matrix $L_G$ of $G$ is defined as follows: 
(i) $L(u,v) = -w_{e_{uv}}$ if $u\neq v$ and (ii)~$L(u,u) = \sum_{v\neq u} w_{e_{uv}}$. 
Let $x$ be an indicator vector for the bipartitioning  
solution $S = \{V_0, V_1\}$ containing 1s in entries corresponding 
to $V_1$, and 0s everywhere else ($V_0$). Then, we have
\begin{equation} \label{eq:cutsize}
  x^T L x = cutsize_G(S).
\end{equation}

\noindent
\textcolor{black}{There is a well-known} connection between 
balanced graph bipartitioning and spectral methods. 
Let $G_C$ be the complete unweighted graph on the vertex set V, 
i.e., for any distinct vertices $u \in V$ and $v \in V$, there exists an 
edge between $u$ and $v$ in $G_C$. Let $L_{G_C}$ denote 
the Laplacian of $G_C$.
Using Equation~(\ref{eq:cutsize}), 
we can express the {\em ratio cut} $R(x)$ \cite{WeiC89} as
\begin{equation}
\label{eq:ratio_cut}
    R(x) \triangleq \frac{cutsize_G(S)}{|S|\cdot|V-S|}
    = \frac{x^T L x}{x^T L_{G_C} x}.
\end{equation}

\noindent
Minimizing $R(x)$ over 0-1 vectors $x$ incentivizes a small $cutsize_G(S)$ 
with a simultaneous balance between $|S|$ and $|V-S|$, hence $R(x)$ can be 
viewed as a proxy for the balanced partitioning objective. 
We relax the minimization problem by looking for real-valued vectors $x$ 
instead of 0-1 vectors $x$,
while ensuring that the real-valued vectors $x$ are orthogonal to 
the common null space of $L$ and $L_{G_C}$~\cite{CucuringuKCMP16}.
A minimizer of Equation (\ref{eq:ratio_cut}) is given by the first 
nontrivial eigenvector of the problem
$L x = \lambda L_{G_C} x$~\cite{CucuringuKCMP16}.

\subsection{Spectral Embeddings and Partitioning} 
\label{subsec:eigen}

A {\em graph embedding} is a map of the vertices in $V$ to points in an $m$-dimensional space. 
In particular, a {\em spectral embedding} can be computed by 
computing $m$ eigenvectors $X\in {\mathbb R}^{|V|\times m}$  of 
a matrix pair $(L_G,B)$, in a generalized eigenvalue problem of the form: 
\begin{equation}
\label{eq:generalized_eigenvalue}
    L_G x = \lambda B x 
\end{equation}
where $L_G$ is a graph Laplacian, 
and $B$ is a positive semi-definite matrix. 
An embedding can be converted into a partitioning by clustering the points in this $m$-dimensional space. 

Spectral embeddings have been used for
hypergraph partitioning. In this context, the hypergraph $H$ is 
first transformed to a graph $G$, and then the spectral embedding is computed using $L_G$.
For example, the eigenvalue problem solved in~\cite{ZienSC99} sets $B=D_w$ 
where $D_w$ is the diagonal matrix containing positive vertex weights. 
In this paper we solve more general problems where $B$ is a graph Laplacian. 
This enables us to handle zero vertex weights as required in practice, 
and to encode in a natural ``graphical'' way prior supervision information 
into the matrix $B$.\footnote{{ \textit{Technical Remark:} In this work, we assume that $G$ is connected. 
Then the problem in Equation (\ref{eq:generalized_eigenvalue}) is well-defined even if $B$ does not correspond to a connected graph, 
because $L_G$'s null space is a subspace of that of $B$ \cite{GhojoghKC19}. The assumption that $G$ is connected holds for practical instances. 
In the more general case we can work by embedding each connected component of $G$ separately and work with a larger embedding. The details are omitted. }}

\subsection{\textcolor{black}{Supervised Dimensionality Reduction (LDA)}} 
\label{subsection:lda}

\textcolor{black}{Linear Discriminant Analysis (LDA) is a supervised algorithm
for dimensionality reduction~\cite{lda}. 
The inputs for LDA are: (i) a matrix $X^{N \times M}$ where the $i^{th}$ row $x_i$ is a point in $M$-dimensional space, and 
(ii) a class label from $\{0,\ldots,K - 1\}$ for each point $x_i$. 
Then, the objective of LDA is to transform $X^{N \times M}$ into $\tilde{X}^{N \times m}$, 
where $m$ ($m<M$) is the target dimension so that the clusters of points corresponding to different classes 
are best separated in the $m$-dimensional space, under the simplifying assumption 
that the classes are normally distributed and class covariances are equal~\cite{lda_tutorial}. 
From an algorithmic point of view, LDA calculates in $O(NM^2)$ time two matrices $S_B^{M\times M}$ 
and $S_W^{M\times M}$ capturing between-class-variance and the within-class-variance respectively. 
Then, it calculates a matrix $P^{M\times m}$ containing the $m$ largest eigenvectors of $S_W^{-1}S_B$, 
and lets $\tilde{X} = X P$. Because in our context $m$ is a small constant, 
LDA can be computed very efficiently.}

\subsection{ILP for Hypergraph Partitioning} 
\label{subsection:ilp}

Hypergraph partitioning can be solved optimally by casting the problem 
as an integer linear program (ILP)~\cite{heuer2015engineering}.
To write balanced hypergraph partitioning as an ILP, for each block $V_i$
we introduce integer \{0,1\} variables, $x_{v,i}$ for each vertex $v$, 
and $y_{e, i}$ for each hyperedge $e$. Setting $x_{v,i}=1$ signifies that vertex $v$ is in block $V_i$, 
and setting  $y_{e,i}=1$ signifies that all  
vertices in hyperedge $e$ are in block $V_i$. 
We then define the following constraints for each $0 \leq i < K$:
\smallskip
\begin{itemize}[noitemsep,topsep=0pt,leftmargin=*]
    \item $\sum_{j = 0}^{K-1}{x_{v,j}} = 1$, for all $v \in V$ 
    \item  $y_{e, i} \leq x_{v, i}$ for all $e \in E$, and $v\in e$
    \item $(\frac{1}{K} - \epsilon) \leq \sum_{v \in V_i}{w_v x_{v, i}} \leq (\frac{1}{K} + \epsilon)W$ \\ where $W= \sum_{v\in V}w_v.$    
\end{itemize}
\smallskip
The objective is to maximize the total weight of the hyperedges that are not cut, i.e.,
$$\textnormal{maximize} \sum_{e \in E}\sum_{i=0}^{K-1}{w_e y_{e ,i}}.$$

\section{\textcolor{black}{The \textit{K-SpecPart} framework}}
\label{sec:frame}

\begin{figure}[!htb]
    \centering
    \includegraphics[width=0.8\columnwidth]{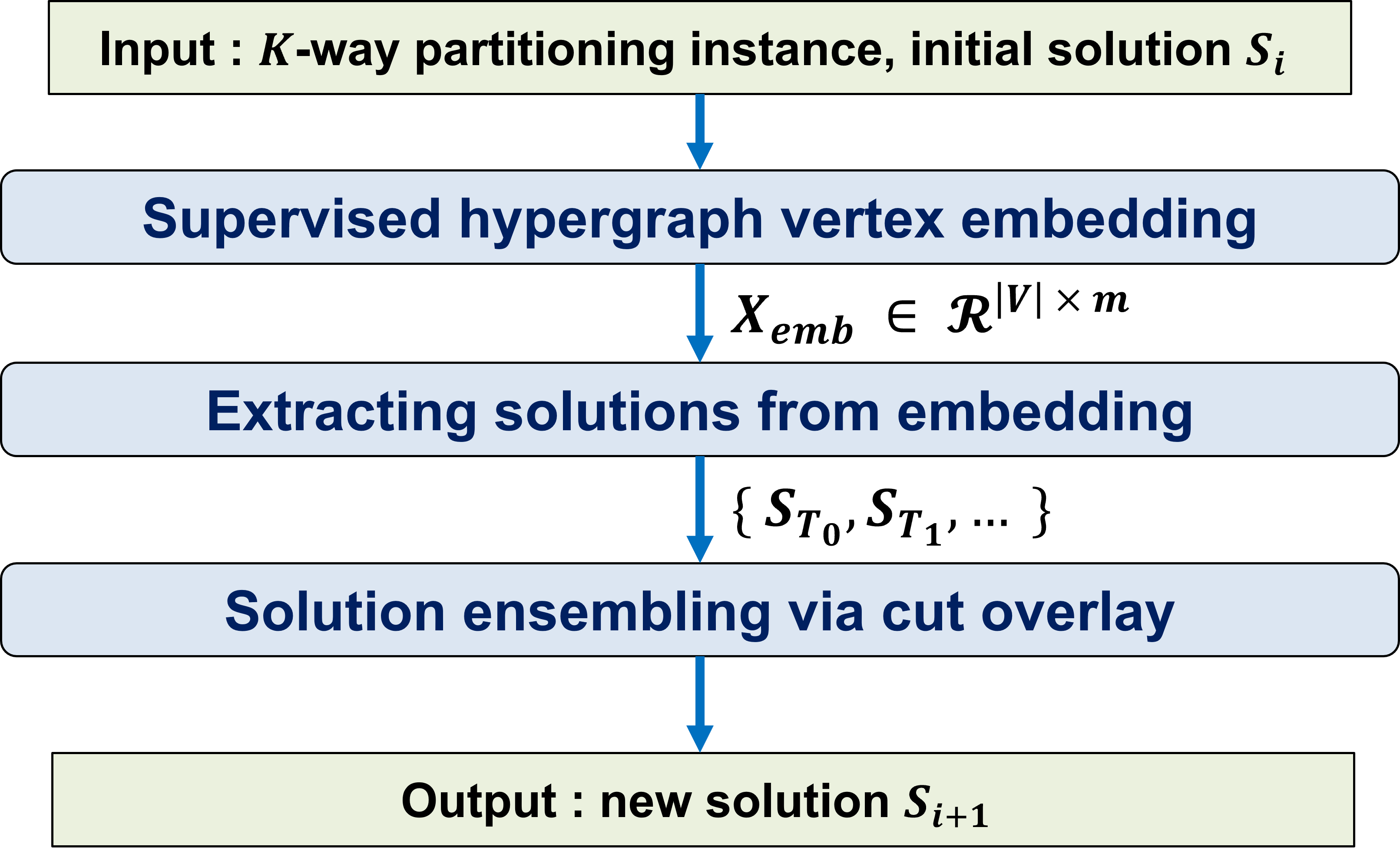}
    \caption{\textcolor{black}{One iteration of {\em K-SpecPart}. The three modules are expanded in 
    Figures~\ref{fig:2_way_embedding},~\ref{fig:hint_generate_k_way} and~\ref{fig:solution_extraction}, 
    and further described in Sections~\ref{sec:embedding}-\ref{sec:overlay_part}. 
    The initial partitioning solution $S_0$ can be
    obtained from any partitioner, but in this work, we use \textit{hMETIS} and \textit{KaHyPar}.  
    During its iterations {\em K-SpecPart} collects its outputs $\{S_0, S_1,\ldots,S_{\beta}\}$. {\em K-SpecPart} 
    then applies the ensembling module on $\{S_0, S_1,\ldots,S_{\beta}\}$ to compute its final output $S_{out}$.}}
    \label{fig:K-SpecPart_frame_}
\end{figure}

\textcolor{black}{We view \textit{K-SpecPart} as an instantiation of a general framework 
for improving a given solution to a partitioning instance. 
The framework involves three modules: {\em vertex embedding module}, 
{\em solution extraction module} and {\em ensembling module},
as illustrated in Figure~\ref{fig:K-SpecPart_frame_}.
The details are given in Algorithm~\ref{alg:k_specpart_pseudocode}.}

\textcolor{black}{The input is the hypergraph $H$ and a partitioning solution $S_i$ 
in the form of block labels $\{0,\ldots,K-1\}$ for the vertices. 
The \textit{vertex embedding module} computes a map of each hypergraph vertex
to a point in a low-dimensional space. 
The embedding is computed by a \textit{supervised} algorithm, using $S_i$ as the supervision input [Alg.~\ref{alg:k_specpart_pseudocode}, Lines 7-19]. 
The intuition is that the vertex embedding is incentivized to conform with $S_i$, 
thus staying in the ``vicinity'' of $S_i$, 
but simultaneously to respect the global structure of the hypergraph, 
thus having the potential to improve $S_i$. 
The {\em solution extraction module} computes a pool of different partitioning 
solutions $\{S_{i,1}\ldots,S_{i,\delta}\}$ [Lines 20-22]. 
These are then sent to the \textit{ensembling module}, 
which uses our cut-overlay method to convert the given solutions 
to a small instance of the $K$-way partition 
which can be solved much more reliably by more expensive partitioning algorithms [Line 23]. 
The solution to this small problem instance is then ``lifted'' (i.e., mapping back to the original hypergraph $H$) and further refined to the output $S_{i+1}$.
The rest of this paper presents our implementations of these three modules.}

\begin{algorithm}[!h]
    \small
    \SetKwData{}{left}\SetKwData{This}{this}\SetKwData{Up}{Up}
    \SetKwInOut{Input}{input}\SetKwInOut{Output}{output}
    \KwInput{Hypergraph $H(V,E)$,  Number of blocks $K$, \\
          \quad \quad \quad Initial partitioning solution $S_{init}$, \\
               \quad \quad \quad Number of supervision iterations $\beta$, \\
               \quad \quad \quad Allowed imbalance between blocks $\epsilon$ \\} 
    \KwOutput{Improved partitioning solution $S_{out}$ \\}
    \BlankLine{}
    Construct the {\em clique expansion graph} $G$ of $H$ and the Laplacian matrix $L_G$ of $G$ (Section \ref{sec:embedding}) \\
    Construct the {\em weight-balance graph} $G_w$ of $H$ and the Laplacian matrix $L_{G_w}$ of $G_w$ (Section \ref{sec:embedding}) \\
    Initialize the empty candidate solution list $\{ S_{candidate} \}$ \\
    $\{ S_{candidate} \}.push\_back(S_{init})$ \\
    $S_0 = S_{init}$ \\
    \For{$i = 0$; $i < \beta$ ; $i++$}{
        \tcc{Supervised hypergraph vertex embedding (Section \ref{sec:embedding})}
        \If{$K = 2$} {
            Construct the {\em hint graph} $G_h$ based on $S_i$ and the Laplacian matrix
            $L_{G_{h_i}}$ of $G_h$ (Section \ref{sec:bipartition}) \\
            Solve the generalized eigenvalue problem 
            $L_G x = \lambda (L_{G_w} + L_{G_{h_i}}) x$ to obtain the first $m$ nontrivial eigenvectors $X_{emb} \in \mathcal{R}^{|V| \times m}$
        } 
        \Else {
            \textcolor{black}{Decompose the $K$-way partitioning solution $S_i$ into $K$ bipartitioning (2-way) solutions $\{S_{b_0}, ..., S_{b_{K-1}}\}$ (Section \ref{sec:kway})\\ 
            \For{$j = 0$; $j < K$; $j++$} {
                Construct the {\em hint graph} $G_{h_j}$ based on $S_{b_j}$ and the Laplacian matrix $L_{G_{h_j}}$ of $G_{h_j}$ (Section \ref{sec:kway}) \\
                Solve the generalized eigenvalue problem $L_{G_j} x = \lambda (L_{G_w} + L_{G_{h_j}}) x$ to obtain the first $m$ nontrivial eigenvectors $X_j^m \in \mathcal{R}^{|V| \times m}$ \\
            }
            $X_{emb} = [X_0^m | X_1^m | ... | X_{K-1}^m]$ \\
            Perform linear discriminant analysis (LDA) to generate the vertex embedding $X_{emb} \in \mathcal{R}^{|V| \times m}$.   }    
        }
        \tcc{Extracting solutions from embedding (Section~\ref{sec:tree_construction})}
        Construct a family of trees $\{T_0, T_1, ...\}$ leveraging the vertex embedding $X_{emb} \in \mathcal{R}^{|V| \times m}$ \\
        Generate hypergraph partitioning solutions $\{S_{T_0}, S_{T_1}, ...\}$ through {\em cut distilling} and tree partitioning \\ 
        \textcolor{black}{Refine $\{S_{T_0}, S_{T_1}, ...\}$ using multi-way FM} \\
        \tcc{Solution ensembling via cut overlay (Section~\ref{sec:overlay_part})}
         \textcolor{black}{$S_{i + 1} \leftarrow$ perform cut-overlay clustering and ILP-based partitioning on the top $\delta$ solutions from $\{S_{T_0}, S_{T_1}, ...\}$} \\     
        $\{ S_{candidate} \}.push\_back(S_{i + 1})$ \\   
    }
    \tcc{Solution ensembling via cut overlay}
    $S_{out} \leftarrow$ perform cut-overlay clustering and ILP-based partitioning on solutions $\{ S_{candidate} \}$   \\      
    \textcolor{black}{Refine $S_{out}$ using multi-way FM} \\
    \Return $S_{out} $ \\   
    \caption{\textcolor{black}{{\em K-SpecPart} framework.}}
    \label{alg:k_specpart_pseudocode}
\end{algorithm}

\section{Supervised Vertex Embedding} 
\label{sec:embedding}

The supervised vertex embedding module takes as \textcolor{black}{inputs} the hypergraph $H(V,E)$ 
and a $K$-way partitioning solution $S_{hint}$,
and outputs an $m$-dimensional embedding $X^{|V|\times m}$.

\begin{figure}[!h]
    \centering
    \includegraphics[width=0.95\columnwidth]{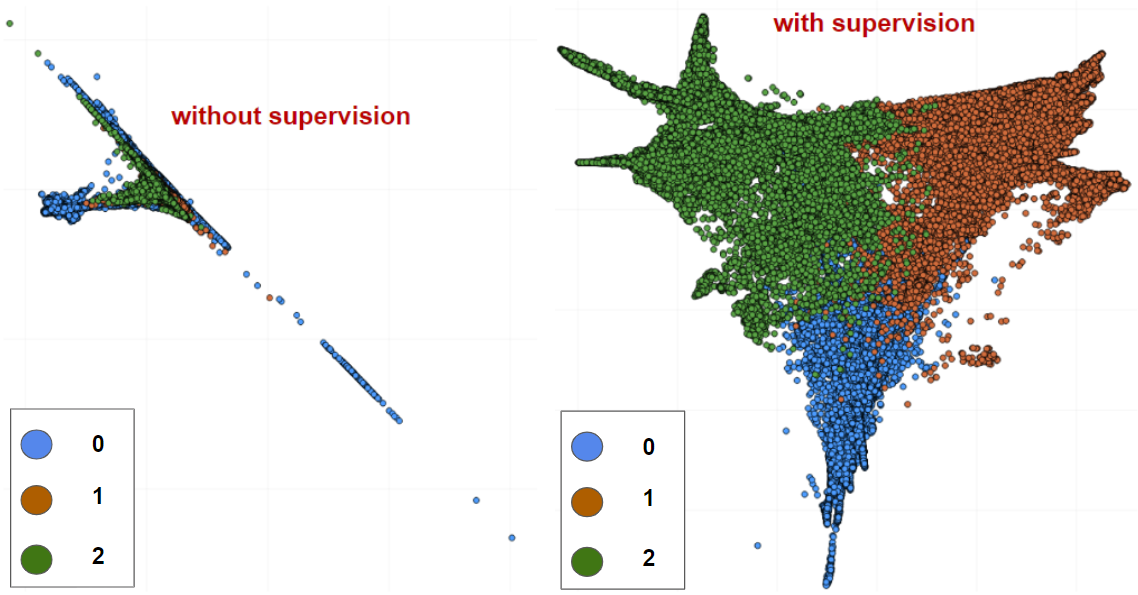}
    \caption{\textcolor{black}{Vertex embeddings of the ISPD IBM14 benchmark. 
    Point colors indicate block membership in a $3$-way partitioning solution 
    with $\epsilon$ $=$ $5\%$ computed by \textit{hMETIS}. 
    The embedding on the right uses as a hint the same \textit{hMETIS} solution, 
    while the embedding on the left is unsupervised.}}
    \label{fig:vertex_embeddings}    
\end{figure}

In \textit{K-SpecPart}, we use a spectral embedding algorithm that 
encodes into a generalized eigenvalue problem the supervision information $S_{hint}$. 
Figure~\ref{fig:vertex_embeddings} illustrates how the inclusion of the 
hint incentivizes the computation of an embedding 
that in general respects (spatially) the given solution $S_{hint}$, 
but also identifies vertices of contention where improving the solution may be possible. 

\subsection{Embedding From Two-way Hint}
\label{sec:bipartition}

\begin{figure}[!t]
    \centering        
    \includegraphics[width=0.75\columnwidth]{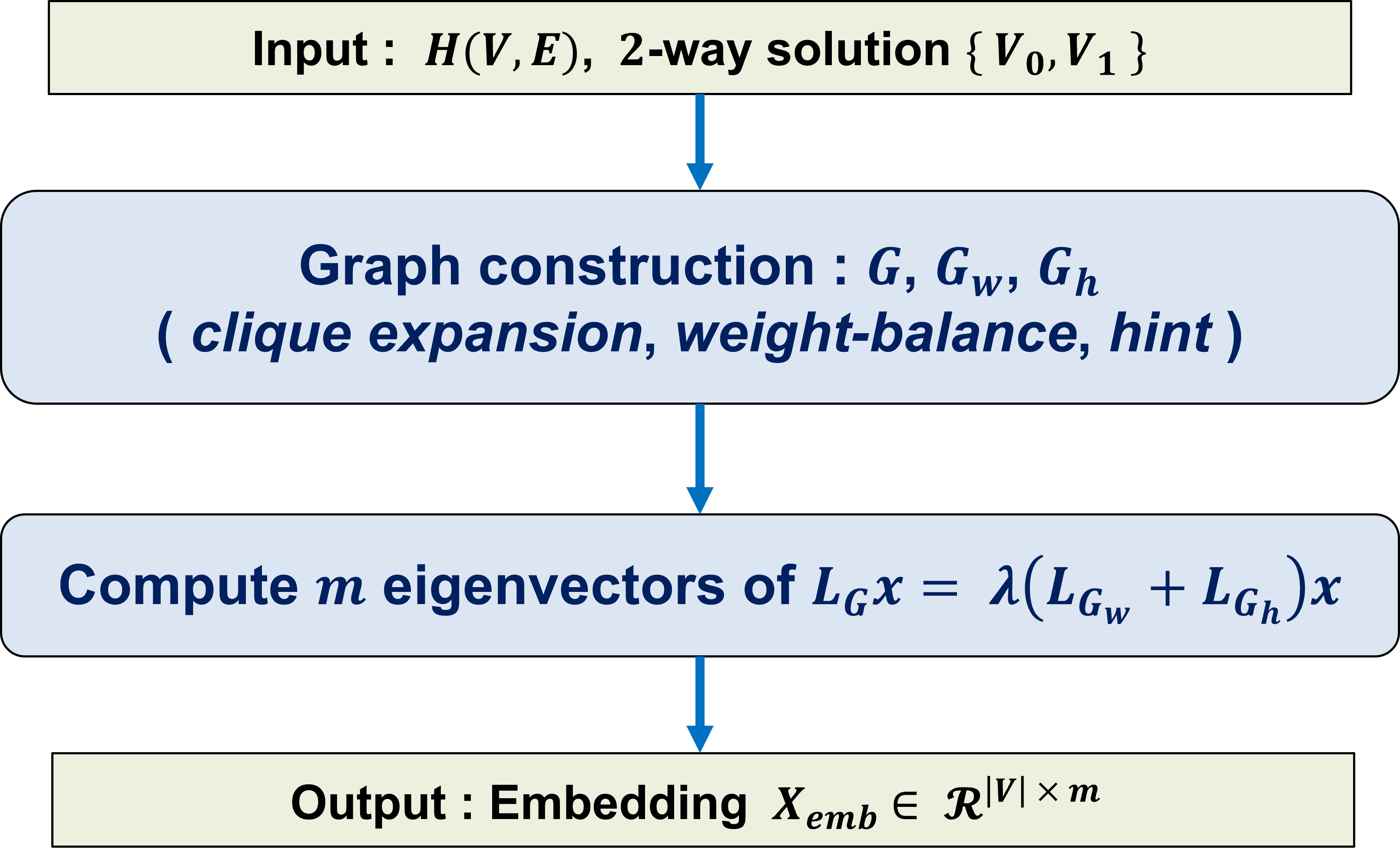}
    \hfill
    \caption{\textcolor{black}{Supervised embedding with a two-way hint.}} 
    \label{fig:2_way_embedding}    
\end{figure}

The embedding algorithm for two-way hints is identical to 
that used in \textit{SpecPart}~\cite{SpecPart}. 
The steps of the algorithm are shown in Figure~\ref{fig:2_way_embedding} 
and described in following paragraphs
that reprise for completeness the corresponding sections in~\cite{SpecPart}. 

\begin{figure}[!htb]
        \centering
        \includegraphics[width=0.87\columnwidth]{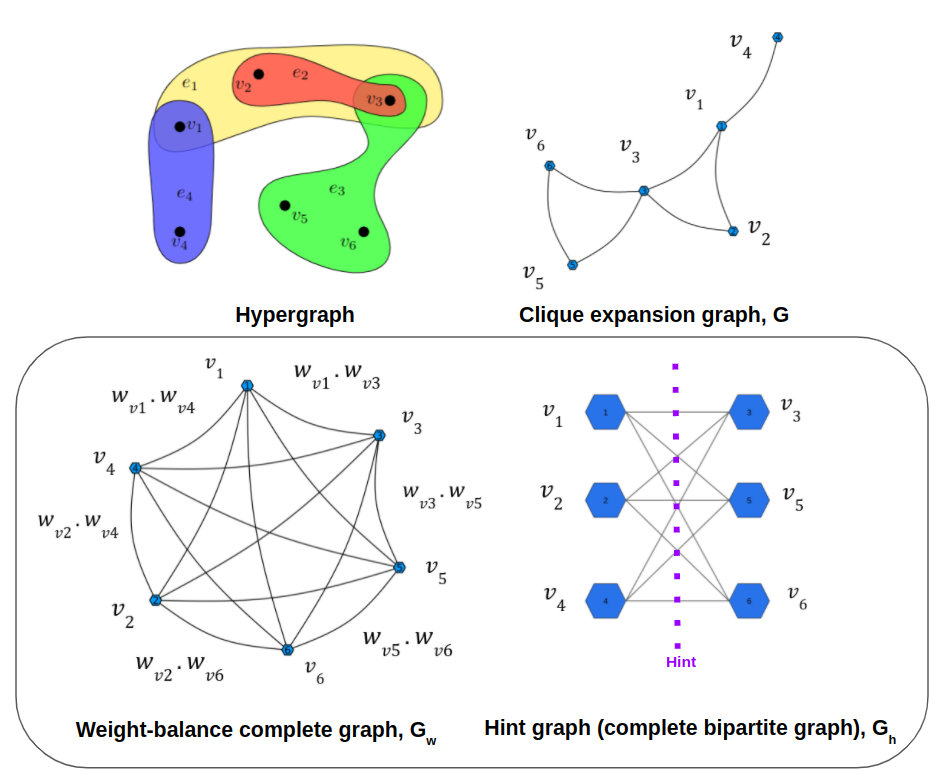}
        \caption{Graphs used for embedding generation. }
        \label{fig:B}
\end{figure}

\vspace{0.25cm}
\noindent
\textbf{Graph Construction.}
We define the graphs used by the embedding algorithm: {\em clique~expansion~graph~$G$},
{\em weight-balance~graph~$G_w$} and {\em hint~graph~$G_h$}.
An illustration of these graphs is given in Figure~\ref{fig:B}.

\smallskip
\noindent $\bullet$ \textit{Clique~Expansion~Graph~$G$:} 
A superposition of weighted cliques. 
The clique corresponding to the hyperedge $e\in E$ has the same 
vertices as $e$ and edge weights  $\frac{1}{|e| - 1}$.
Graph $G$ has size $\sum_{e\in E}{\frac{|e|(|e| - 1)}{2}}$ where $|e|$ is 
the size of hyperedge $e$. This is usually quite large relative to the input 
size $|I| = \sum_{e\in E} |e|$. For this reason, we only construct a function 
$f_{L_G}$ that evaluates matrix-vector products of the form $L_Gx$, where 
$L_G$ is the Laplacian of $G$, which is all we need to perform the 
eigenvector computation. In all places where we mention the construction of 
any Laplacian, we construct the equivalent function for evaluating 
matrix-vector products. The function $f_{L_G}$ is an application of the following equation that is based 
on expressing $L_G$ as a sum of Laplacians of cliques:
\begin{equation}
\resizebox{0.35\textwidth}{!}{
   $L_G x = \sum_{e\in E} \frac{1}{|e|-1}\left(x - \frac{x^T{\bf 1_e}}{{\bf 1_e}^T {\bf 1_e}}\cdot {\bf 1_e}\right)$}
\end{equation}
where ${\bf 1_e}$ is the 1-0 vector with 1s in the entries corresponding to 
the vertices in $e$. By exploiting the sparsity in ${\bf 1_e}$, the product 
is implemented to run in $O(|I|)$ time.

\smallskip

\noindent
\textit{$\bullet$ Weight-Balance~Graph~$G_w$}: A complete weighted graph used 
to capture arbitrary vertex 
weights and incentive balanced cuts.
$G_w$ has the same vertices as hypergraph $H$, 
and edges of weight $w_u \cdot w_v$ between any two vertices $u$ and $v$. 
Let $W_{V_i}$ be the weight of block $V_i$, i.e.,
$W_{V_i} = \sum_{v \in V_i}{w_v}.$
Then, given a two-way solution $S_b=\{V_0,V_1\}$, we have 

\begin{equation} \label{eq:Bcutsize}
\begin{split}
    W_{V_0} \cdot W_{V_1} 
        &= \sum_{v \in V_0}{w_v} \cdot \sum_{u \in V_1}{w_u} = \sum_{v \in V_0, u \in V_1}{ w_v \cdot w_u} \\
        &=\sum_{v \in V_0, u \in V_1}{ w_{e_{vu}}    } 
                           = cutsize_{G_w}(S_b).
\end{split}
\end{equation}

We now discuss how to compute matrix-vector products with the Laplacian 
matrix of $G_w$. Let ${\bf w}$ be the vector 
of vertex weights. We apply the identity
\begin{equation}
    L_{G_w} x = {\bf w} \circ x - \frac{x^T {\bf 1}}{1^T \bf 1}\cdot {\bf w}
\end{equation}
where ${\bf 1}$ is the all-ones vector and $\circ$ denotes the Hadamard product. This can be carried out in time $O(|V|)$.

In general, any vector $x$ can be written in the form $x = y + c{\bf 1}$, 
where $y^T {\bf 1}=0$. Substituting this decomposition of $x$ into the above 
equation, we get that $L_{G_w}x = {\bf w} \circ y.$
In other words, $L_{G_w}$ acts like a diagonal matrix on $y$ and nullifies the constant component of $x$.

\smallskip
\noindent
\textit{$\bullet$ Hint~Graph~$G_h$:} A complete bipartite graph on the two vertex sets  $V_0$ and $V_1$ 
defined by the two-way hint solution $S_{{b}}$.  We have

\begin{equation}
    \begin{split}
        L_{G_h} x &= (x- \frac{x^T{\bf 1}}{1^T 1}\cdot{\bf 1}) 
                - (x- \frac{x^T{\bf 1_{V_0}}}{1_{V_0}^T 1_{V_0}}\cdot{\bf 1_{V_0}}) \\
                &-(x- \frac{x^T{\bf 1_{V_1}}}{1_{V_1}^T 1_{V_1}}\cdot{\bf 1_{V_1}}) 
    \end{split}
\end{equation}

\noindent
where ${\bf 1}_{V_i}$ denotes the 1-0 vector with 1s in entries 
corresponding to the vertices in $V_i$. By exploiting the sparsity 
in ${\bf 1}_{V_i}$, the product is implemented in $O(|V|)$ time. 

\vspace{0.25cm}
\noindent
\textbf{Generalized Eigenvalue Problem and Embedding.}
Given a two-way partitioning solution $S_{hint}$, we solve the generalized eigenvalue problem $L_G x=  \lambda B x$ 
where $B = L_{G_w}+L_{G_h}$, and compute the first $m$ nontrivial eigenvectors $X\in {\mathbb R}^{|V|\times m}$ 
whose rows provide the vertex embedding. 

From the discussion in  
Section~\ref{sec:rq} recall that
the eigenvalue problem is directly related to solving
\begin{equation}
\min_x R(x)  = \min_x  \frac{x^T L_G x}{x^T L_{G_w} x + x^T L_{G_h} x }
\label{eq:gen-eigen}
\end{equation}

\noindent
over the real vectors $x$. Recall also that this is a relaxation of the  
problem over 0-1 indicator vectors. Let $x_S$ be the indicator 
vector for some set $S\subset V$. Then, using Equation~(\ref{eq:cutsize}) 
we can understand the rationale for Equation (\ref{eq:gen-eigen}): 
\begin{itemize}[noitemsep,topsep=3pt,leftmargin=*]
    \item $x_S^T L_G x_S = cutsize_G(S)$ which is \textcolor{black}{a} proxy 
    for $cutsize_H(S_b)$. Thus, the {\em numerator} 
    incentivizes {\em smaller} cuts in $H$. 
    \item $x_S^T L_{G_w} x_S = cutsize_{G_w}(S)$. By 
    Equation~(\ref{eq:Bcutsize}), this is equal to $W_S\cdot W_{V-S}$, where $W_S$ is the total weight of the vertices in $S$.  
    Thus, the {\em denominator} incentivizes a {\em large} 
    $W_S\cdot W_{V-S}$, which implies balance. 
    \item $x_S^T L_{G_h} x_S$ is maximized when all edges of $G_{h}$ are cut. Thus, the {\em denominator} incentivizes 
    cutting {\em many} edges that are also cut by the hint. 
\end{itemize}

\vspace{0.25cm}
\noindent
\textbf{Generalized Eigenvector Computation.}
We solve $L_G x = \lambda B x$ 
using \textit{LOBPCG}, an iterative preconditioned
eigensolver. \textit{LOBPCG} relies on functions that evaluate matrix-vector products with $L_G$ and $B$. For fast
computation, the solver can utilize a preconditioner for $L_G$, also in
an implicit functional form. To compute the preconditioner we first obtain
an explicit graph $\tilde{G}$ that is spectrally similar with $G$ and
has size at most \textcolor{black}{$2|I|$}, where $|I| = \sum_{e \in E} |e|$.  
More specifically,
we build $\tilde{G}$ by replacing every hyperedge $e$ in $H$ with the sum
of \textcolor{black}{2} uniformly weighted {\em random cycles} on the vertices $V_e$ of~$e$.
This is an essentially optimal sparse spectral approximation 
for the clique on $V_e$, as implied from asymptotic properties of random 
$d$-regular expanders~(e.g.,~see~\cite{KapralovP12}
or Theorem 4.16 in~\cite{HorryL06}).
Since $G$ is a sum of cliques, and $\hat{G}$ is a sum of tight spectral
approximations of cliques, graph support theory~\cite{BomanH03}
implies that $\hat{G}$ is a tight spectral approximation for $G$. 
Finally, we compute a preconditioner of $L_{\hat{G}}$ using the \textit{CMG}
algorithm~\cite{KoutisMT11} and in particular the implementation from~\cite{Julia-CMG}\textcolor{black}{.}
By transitivity~\cite{BomanH03} the preconditioner for $L_{\hat{G}}$  is also a preconditioner for $L_G$.

\subsection{\textcolor{black}{Embedding From Multi-way Hint}}
\label{sec:kway}

\begin{figure}[!htb]
    \centering
    \includegraphics[width=0.995\columnwidth]{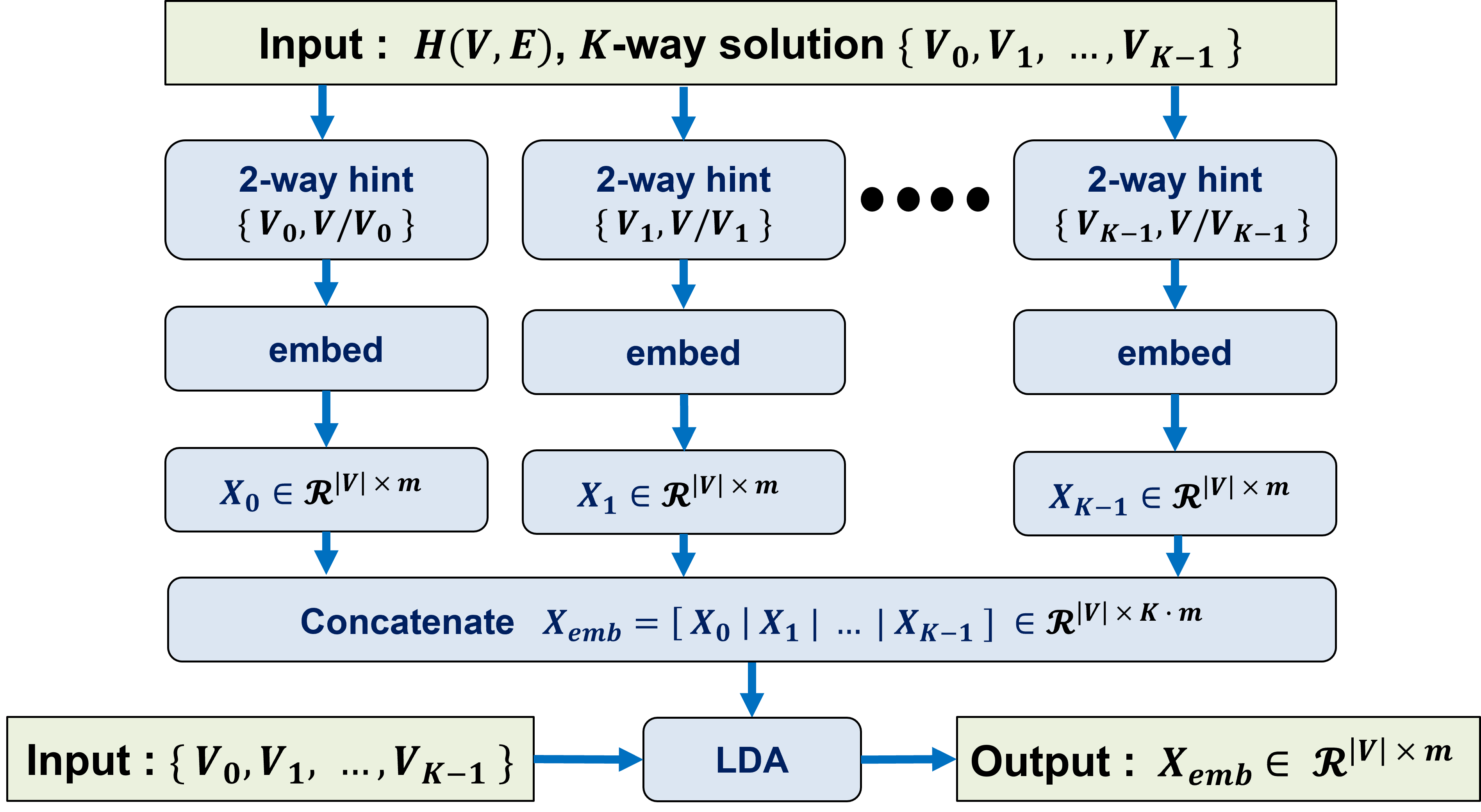}
    \caption{\textcolor{black}{Vertex embedding generation process for a given multi-way (\textit{$K$ $>$ $2$}) hint, 
              using the two-way embedding subroutine from Section~\ref{sec:bipartition}}.}
    \label{fig:hint_generate_k_way}
\end{figure}

\textcolor{black}{The flow for generating an $m$-dimensional embedding from a $K$-way hint is shown in Figure~\ref{fig:hint_generate_k_way}.
The steps are described in \textcolor{black}{the} following paragraphs.}

\noindent
\textcolor{black}{\textbf{Embedding by concatenation.}
In the $K$-way case where $K>2$, the solution hint $S_{hint}$ corresponds to
a $K$-way partitioning solution $\{V_0,\ldots,V_{K-1}\}$. 
We then extract $K$ different bipartitions, $S_{b_i}$, for $i=0,\ldots,K-1$, where
$$
     S_{b_j} = \{V_j, \bigcup_{i = 0, i \neq j}^{K-1}V_i\}.
$$
For each $S_{b_j}$ we solve an instance of the generalized problem 
we set up in Section~\ref{sec:bipartition}. 
This generates $K$ different embeddings $X_{j} \in {\mathbb R}^{|V|\times m}$. 
We then concatenate these $K$ embeddings horizontally to get our final 
embedding $X_{emb} \in {\mathbb R}^{|V|\times {K\cdot m}}$, i.e.,
$$
     X_{emb} = [ X_0 | X_1 | \ldots | X_{K-1} ] , ~\textnormal{where~} X_j \in {\mathbb R}^{|V|\times m}.
$$}

\noindent
\textcolor{black}{\textbf{Supervised Dimensionality Reduction.}
Note that the above embedding $X_{emb}$ has dimension $K\cdot m$. 
We then apply on $X_{emb}$ a supervised dimensionality reduction algorithm, 
specifically LDA (see Section \ref{sec:lda}), as illustrated in Figure~\ref{fig:hint_generate_k_way}.
We use LDA primarily to reduce the runtime of subsequent steps, 
but also because \textcolor{black}{this} second application of supervision has the potential to increase the quality of the embedding. }

\textcolor{black}{Besides $X_{emb}$, LDA takes as input a target dimension, 
and class labels for the points in $X_{emb}$. 
We choose $m$ as the target dimension. 
We assign label $i$ to vertex $v$ if $V\in V_i$. 
For the computation, we use a Julia-based LDA implementation 
from the {\em MultivariateStats.jl} package \cite{ldapack}.}

\section{Extracting Solutions from Embeddings}
\label{sec:tree_construction}

\begin{figure}[!htb]
    \centering
    \includegraphics[width=0.85\columnwidth]{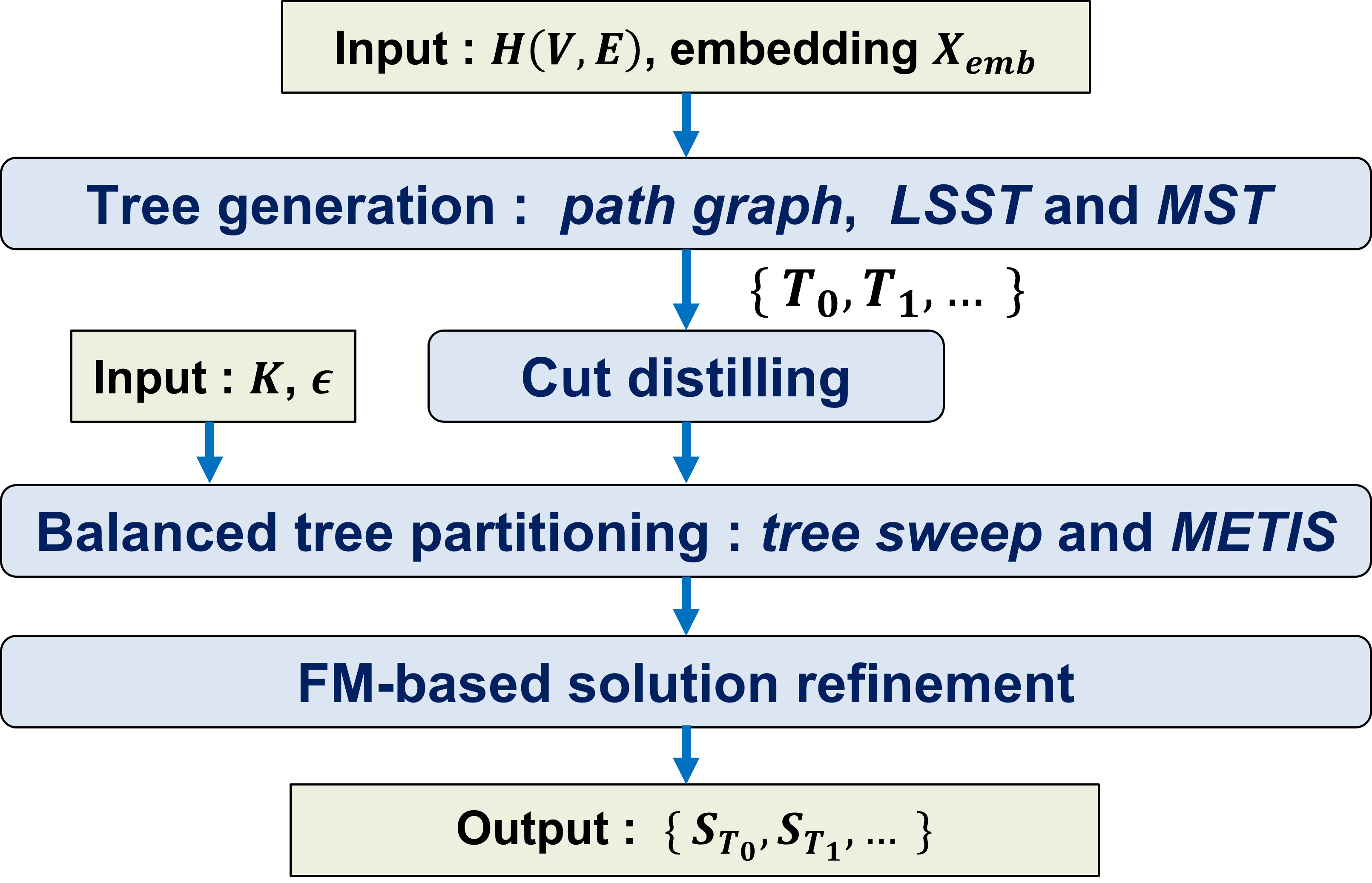}
    \caption{\textcolor{black}{The flow of extracting solutions from embeddings.}}
    \label{fig:solution_extraction}
\end{figure}

The inputs of \textcolor{black}{the} solution extraction module are 
the hypergraph $H$, number of blocks $K$, 
balance constraint $\epsilon$ and an embedding
$X_{emb} \in \mathcal{R}^{|V| \times m}$, 
and the output is a pool of solutions $\{S_0,\ldots,S_{\delta-1}\}$. 
The main idea of the algorithm is to use the embedding to reduce the $K$-way hypergraph partitioning problem to multiple $K$-way balanced partitioning problems on trees whose edge weights ``summarize'' the underlying cuts of the hypergraph. 
The steps of the algorithm are shown in Figure~\ref{fig:solution_extraction} and described 
in Sections~\ref{sec:tree_partitioning} \textcolor{black}{and} \ref{sec:cut_distilling}.

\subsection{Tree Generation}
\label{sec:treegeneration}
In our algorithm, each $S_i$ in the output comes from a tree that spans the set of vertices $V$. Here we define the types of trees we use. 

\smallskip 
\noindent
\textbf{Path Graph.}
We first define a path graph on the vertices $V$, which appears in the proofs of Cheeger inequalities for bipartitioning~\cite{Chung1997,KMP23}. Let $X_{emb_i}$ be the $i^{th}$ column of $X$. We sort the values in $X_{emb_i}$ and let $o(j)$ be vertex at the $j^{th}$ position of the sorted $X_{emb_i}$. Then we define the path graph on $V$ to be $v_{o(1)}, v_{o(2)},\ldots,v_{o(|V|)}.$

\smallskip
\noindent
\textbf{Clique Expansion Spanning Tree.}
The path graph is likely not a spanning tree of the clique expansion graph $G$. To take connectivity directly into account, we work with a weighted 
graph that reflects both the connectivity of~$H$ and the global information contained in the embedding, adapting an idea that has been used in work on $K$-way Cheeger inequalities~\cite{lee2014multiway}. 
Concretely, we form a graph $\hat{G}$ by replacing every hyperedge $e$ of $H$ with a sum of $\zeta$ cycles~(as also done in Section~\ref{sec:bipartition}). Suppose that $Y \in {\mathbb R}^{|V|\times 
d}$ is an embedding matrix. We denote by \textcolor{black}{$Y_u$} the row of~$Y$ containing the embedding of vertex $u$. We construct 
the weighted graph $\hat{G}_Y$ by setting the weight of each 
edge $e_{uv}\in \hat{G}$ to $||Y_u - Y_v||_2$, i.e.,~equal to the 
Euclidean distance between the two vertices in the embedding. Using $\hat{G}_Y$ we build two spanning trees.

\noindent
$\bullet$ \textit{LSST:} A desired property for a spanning tree  
$\hat{T}$ of $\hat{G}_Y$ is to preserve the embedding information 
contained in $\hat{G}$ as faithfully as possible. Thus, we let 
$\hat{T}$ be a {\em Low Stretch Spanning  Tree (LSST)} of~$\hat{G}$, which by 
definition means that the weight $w_{e_{uv}}$ of each edge in  $\hat{G}$ is  
approximated {\em on average}, and up to a small function $f(|V|)$, by the 
distance between the nodes $u$ and $v$ in $\hat{T}$~\cite{AlonKPW95}. We 
compute the LSST using the AKPW algorithm of Alon et al.~\cite{AlonKPW95}. 
The output of the AKPW algorithm depends on the vertex ordering of its 
input. To make it invariant to the vertex ordering in the original 
hypergraph~$H$, we relabel the vertices of $\hat{G}_Y$ using the order induced by sorting 
the smallest nontrivial eigenvector computed earlier. Empirically, this 
order has the advantage of producing LSSTs that \textcolor{black}{contain} slightly better cutsizes.

\noindent
$\bullet$ \textit{MST:} A graph can contain multiple different LSSTs, with each of 
them approximating to different degrees the weight $w_{e_{uv}}$ for any given 
$e_{uv}$. It is known that the AKPW algorithm is  
suboptimal with respect to the approximation factor~$f(|V|)$; more 
sophisticated algorithms exist but they are far from practical. 
Hence, we also apply Kruskal's algorithm~\cite{Kruskal56} to compute a 
Minimum Spanning Tree of~$\hat{G}$, which serves as 
an easy-to-compute proxy to an LSST. 
The MST can potentially have better or complementary distance-preserving 
properties relative to the tree computed by the AKPW algorithm. 

In summary, we compute $m$ path graphs, and also generate the LSSTs and MSTs by letting $Y$ range over each subset of columns of $X_{emb}$. This produces a family of $t=2(2^m-1)+m$ trees.

\subsection{Cut Distilling}
\label{sec:cut_distilling}
We reweight {\em each} tree $T$ in the given family of trees to distill the cut structure of $H$ over $T$, in the following sense. 
(i) For a given tree $T=(V,E_T)$, observe that the removal of an edge $e_T$ of $T$ yields a partitioning  $S_{e_T}$ of $V$ and thus of 
the original hypergraph $H$. (ii) We reweight each edge $e_T\in E_T$ with the corresponding $cutsize_H(S_{e_T})$.

With this choice of weights, we have $cutSize_H(S) \leq cutSize_T(S)$, and owing to the reasoning behind the construction of $T$, $cutSize_T(S)$ provides a proxy for $cutSize_H(S)$. 

\begin{figure}[!htb]
    \centering
    \includegraphics[width=0.75\columnwidth]{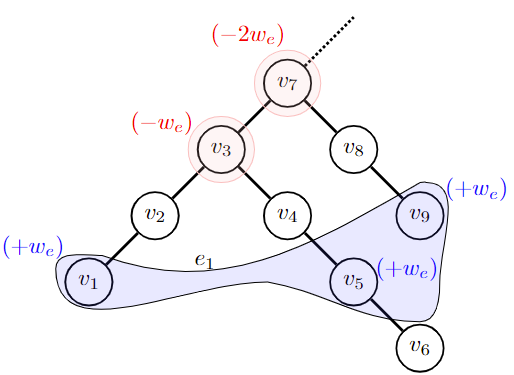}
    \caption{Hyperedge, junctions and their numerical labels. The vertices highlighted in red are the junction vertices. }. 
    \label{fig:treesweep}
\end{figure}

Computing edge weights on $T$ can be done in $O(\sum_e |e|)$ time, 
via an algorithm involving the computation of least common 
ancestors (LCA) on $T$, in combination with dynamic programming on $T$~\cite{bender2000lca}. \textcolor{black}{We 
provide pseudocode in Algorithm~\ref{alg:cut_distilling} and give a fast implementation in \cite{TILOS-HGPart}.} We illustrate the idea using the example in Figure~\ref{fig:treesweep}.

We consider $T$ to be rooted at an arbitrary vertex.
In the example of Figure~\ref{fig:treesweep}, consider hyperedge $e= \{v_1, v_5, v_9\}$. 
The LCA of its vertices is~$v_7$. 
Then, the weight of $e$ should be accounted for the set $C_e \subset E_T$ of 
all tree edges that are ancestors of $\{v_1, v_5, v_9\}$ and descendants 
of~$v_7$. We do this as follows \textcolor{black}{[Alg.~\ref{alg:cut_distilling}, Lines 2-13]}.
(i) We compute a set of {\em junction} vertices
that are LCAs of $\{v_1,v_5\}$ and $\{v_1,v_5,v_9\}$. 
(ii) We then ``label'' these junctions with $-w_e$, where $w_e$ is the 
weight of $e$.
More generally, for a hyperedge $e=\{v_{i_1},\ldots,v_{i_p}\}$ ordered 
according to the post-order depth-first search traversal on $T$, 
we calculate the LCAs for the $p-1$ sets $\{v_{i_1},\ldots,v_{i_j}\}$ for 
$j=2,\ldots,p$, 
and the junctions are labeled with appropriate negative multiples of $w_e$. 
We also label the vertices in $e$ with $w_e$. (iii) All other vertices are 
labeled with 0. 

\begin{algorithm}[!h]
    \small
    \SetKwData{}{left}\SetKwData{This}{this}\SetKwData{Up}{Up}
    \SetKwInOut{Input}{input}\SetKwInOut{Output}{output}
    \KwInput{\textcolor{black}{Hypergraph $H(V,E)$, Tree $T(V, E_T)$}} 
    \KwOutput{\textcolor{black}{Tree $T$ with updated edge weights \\}}
    \BlankLine{}
    {
    \textcolor{black}{Select an arbitrary vertex $v_{root}$ from $V$ 
    and root $T$ at $v_{root}$ \\
    Perform a post-order depth-first search traversal on $T$
    and store the sequence of visited vertices in $visited\_sequence$ \\
    \tcc{Label each vertex in $T$ based on the hyperedge weight $w_e$
    for $e \in E$}
    $cuts\_{delta}[v] \leftarrow 0$ for all $v$ in $V$ \\ 
    \For{each $e$ in $E$}{
        $cuts\_{delta}[v] = cuts\_{delta}[v] + w_e$ for all $v$ in $e$ \\
        $\{v_{i0}, ..., v_{i(|e| - 1)} \} \leftarrow$ arrange the vertices
        of $e$ according to their positions in $visited\_sequence$ \\
        $v_{LCA} \leftarrow$ $v_{i0}$ \\
        \For{$j = 1$; $j < |e|$; $j++$} {
            $v_{LCA} \leftarrow$ identify the least common ancestor (LCA) for $v_{LCA}$ and $v_{ij}$ in $T$ \\
            $cuts\_{delta}[v_{LCA}]$ = $cuts\_{delta}[v_{LCA}] - w_e$ \\
        }
        $cuts\_{delta}[v_{LCA}]$ = $cuts\_{delta}[v_{LCA}] - w_e$ \\
    }
    \tcc{Reweight edges of $T$}
    \For{each $e_T$ in $E_T$} {
        $w_{e_T} \leftarrow$ compute the sum-below-$e_T$ (i.e., the sum of the labels $cuts\_{delta}$ of vertices  that are descendants
        of $e_T$) in the post-order depth-first search ordering
        \\
    }        
    \Return $T$ with updated edge weights \\  
    }}
    \caption{\textcolor{black}{Cut Distilling}}
    \label{alg:cut_distilling}
\end{algorithm}

Consider then an arbitrary edge $e_T$ of the tree, and compute 
the sum-below-$e_T$, i.e.,~the sum of the labels of vertices that are
{\em  descendants} of $e_T$. This will be $w_e$ on all edges of $C_e$ and 0 
otherwise, thus correctly accounting for the hyperedge $e$ on the intended 
set of edges $C_e$ \textcolor{black}{[Alg.~\ref{alg:cut_distilling}, Lines 14-16]}. In order to compute the correct total 
counts of cut hyperedges on all tree edges, we iterate over hyperedges, 
compute their junction vertices, and aggregate the associated labels. Then, 
for any tree edge $e_T$, the sum-below-$e_T$ will equal 
$cutsize_H(S_{e_T})$. These sums can be computed in $O(|V|)$ time, via 
dynamic programming on $T$.

\subsection{Tree Partitioning}
\label{sec:tree_partitioning}
We use a linear ``tree-sweep'' method and {\em METIS} to partition the trees.
\textcolor{black}{In our studies, we have observed \textcolor{black}{that} only using {\em METIS} as 
the tree partitioner results in an average of 3\%, 4\% and 3\% 
deterioration in cutsize for $K=$ 2, 3 and 4 respectively.}

\noindent $\bullet$ \textit{$K$ $=$ $2$}. Given a cut-distilling tree $T$, and referring back to Figure~\ref{fig:treesweep}, an application of dynamic programming can compute the total weight of the vertices that lie below $e_T$ on $T$. We  can thus compute the value for the balanced cut objective for $S_{e_T}$ and  pick the $S_{e_T}$ that minimizes the objective among the $n-1$ cuts  suggested by the tree. This ``tree-sweep'' algorithm generates a good-quality two-way partitioning 
solution from the tree. 
Additionally, 
we use {\em METIS}~\cite{KarypisK98} to solve a balanced two-way 
partitioning problem on the edge-weighted tree, with the original vertex 
weights from~$H$. In some cases, this improves the solution. 

\noindent
\textcolor{black}{$\bullet$ \textit{$K$ $>$ $2$.} Similar to $K=2$, we use two algorithms to compute two potentially different $K$-way partitioning solutions of the tree. The first algorithm is \textit{METIS}~\cite{KarypisK98}. 
The second algorithm extends the two-way cut partitioning of 
the tree to $K$-way partitioning. To this end, we apply the 
two-way algorithm recursively, for $K$ $-$ $1$ levels.
We use a similar idea as the VILE (``very illegal'') method 
\cite{Caldwell2000} to generate an imbalanced partitioning solution and then 
refine the solution with the FM algorithm.
Specifically, while computing the $i^{th}$ 
level bipartitioning solution $S_{e_T}^i$ on the tree $e_T$, the balance constraint for block $V_{i0}$ in the 
bipartitioning solution $S(V_{i0}, V_{i1})$ 
is: $(\frac{1}{K}- \epsilon)\cdot W  \leq  \sum_{v \in V_{i0}}{w_v}  
\leq (\frac{1}{K}+ \epsilon) \cdot W$}.

\textcolor{black}{After obtaining the $i^{th}$ bipartitioning solution $S(V_{i0}, V_{i1})$,
we mark all the vertices in $V_{i0}$ as fixed vertices and 
set their weights to zero. 
We then proceed with the $(i + 1)^{th}$ level bipartitioning solution $S_{e_T}^{i+1}$
on the tree $e_T$.}

\subsection{\textcolor{black}{Refinement on the hypergraph}}
\label{sec:FMrefinement}

\textcolor{black}{The previous step solves balanced partitioning on trees that share the same vertex set $V$ with $H$. 
Note that the number of solutions will be larger than the number of trees $t$, 
because we apply different partitioning algorithms \textcolor{black}{to} each tree. 
These solutions are then transferred to $H$, and \textcolor{black}{each is} further refined using the FM algorithm~\cite{FiducciaM82} on the entire hypergraph $H$. 
In particular, we use the FM implementation in~\cite{TritonPart}.}

\section{Solution ensembling via cut overlay}
\label{sec:overlay_part}
The input of this module is the given $K$-way partitioning instance 
and a pool of partitioning solutions.
We then perform the following steps.

\noindent
\textbf{Cut-Overlay Clustering.}
We first select the $\delta$ best solutions. 
Let $E_{1},\ldots, E_{\delta} \subset E$ be 
the sets of hyperedges cut in the $\delta$ solutions. We remove 
the union of these sets from $H$ to yield a number of connected clusters. 
Then, we perform a cluster contraction process that is standard in 
multilevel partitioners, to give rise to a clustered hypergraph 
$H_c(V_c,E_c)$. By construction, $E_c$ consists of $E_1 \cup 
... \cup E_{\delta}$ and hence is guaranteed to contain a solution which is 
{\em at least as good} as the best among the cuts $E_i$.

\noindent
\textbf{ILP-based Partitioning.}
The coarse 
hypergraph ($H_c$) obtained from \textcolor{black}{cut-overlay} clustering usually \textcolor{black}{has} a few hundreds 
of vertices and hyperedges (\textcolor{black}{including with} the default setting $\delta = 5$). 
While even this small size would be expected to be prohibitive for applying an exact optimization algorithm, somewhat surprisingly, an ILP formulation can frequently solve the problem optimally. 
In most cases, our ILP produces a solution better than any of the $\delta$
candidate solutions. 
We solve the ILP with the CPLEX solver~\cite{CPLEX}. We have found that the open-source OR-Tools package~\cite{ortools} is significantly slower. 
In our current implementation, we include a parameter 
$\gamma$: in the case when the number of hyperedges in $H_c$ is larger than 
$\gamma$, we run {\em hMETIS} on $H_c$. This step generates a  $K$-way solution $S'$ on $H_c$.

\noindent
\textcolor{black}{\textbf{Lifting and Refinement.}}
\textcolor{black}{The solution $S'$ from the previous step is ``lifted'' to $H$, with the standard lifting process that multilevel partitioners use. Finally, we apply FM refinement on $H$ to obtain the final solution $S$. Here we \textcolor{black}{again use} the FM implementation from~\cite{TritonPart}. }

\section{Experimental Validation}
\label{sec:experiment}

The {\em K-SpecPart} framework is implemented in Julia. We use {\em CPLEX}~\cite{CPLEX} and  {\em LOBPCG}~\cite{KnyazevLAO07} 
as our ILP solver (we provide an OR-Tools based implementation) and eigenvalue 
solver respectively. We run all experiments on a server with an Intel Xeon E5-2650L, 1.70GHz CPU and 256 GB memory. 
We have compared our framework with two state-of-the-art hypergraph partitioners
({\em hMETIS}~\cite{KarypisAKS99} and 
{\em KaHyPar}~\cite{SebastianTLYCP22}) 
on the {\em ISPD98 VLSI Circuit Benchmark Suite}~\cite{Alpert98} 
and the {\em Titan23 Suite}~\cite{MurrayWLLB13}. We make 
public all partitioning solutions, scripts and 
code at \cite{TILOS-HGPart}.

\subsection{\textcolor{black}{Cutsize Comparison}}
\label{sec:experiments}

\textcolor{black}{We run {\em hMETIS} and {\em KaHyPar} with their 
respective default parameter 
settings.\footnote{The default parameter setting 
for {\em hMETIS}~\cite{hMETISManual}
is: Nruns = 10, CType = 1, RType = 1, Vcycle = 1,  Reconst = 0 and seed = 0.
The default configuration file we use for KaHyPar 
is cut\_rKaHyPar\_sea20.ini~\cite{KaHyParConfig}.} We denote by \textit{hMETIS}$_t$, the best cutsize obtained by $t$ runs of \textit{hMETIS},  with $t$ different seeds. We denote by \textit{hMETIS}$_{avg}$, the average (over 50 samples) cutsize of \textit{hMETIS}$_{20}$. We adopt similar notation for \textit{KaHyPar}. In all our experiments we run {\em K-SpecPart} with its default settings (see  Table~\ref{tab:term}) and a hint that comes from  {\em hMETIS}$_1$. We compare {\em K-SpecPart} against {\em hMETIS}$_5$ and {\em KaHyPar}$_5$; this is because 5 runs of \textit{hMETIS} have a similar runtime with \textit{K-SpecPart}. For a more robust and challenging comparison, we also compare {\em K-SpecPart} against {\em hMETIS}$_{avg}$ and {\em KaHyPar}$_{avg}$, which gives to these partitioners at least $\sim$4X the walltime of \textit{K-SpecPart}.}
\begin{figure}[!t]
    \centering
    \begin{subfigure}[b]{0.48\textwidth}
        \centering
        \includegraphics[width=\textwidth]{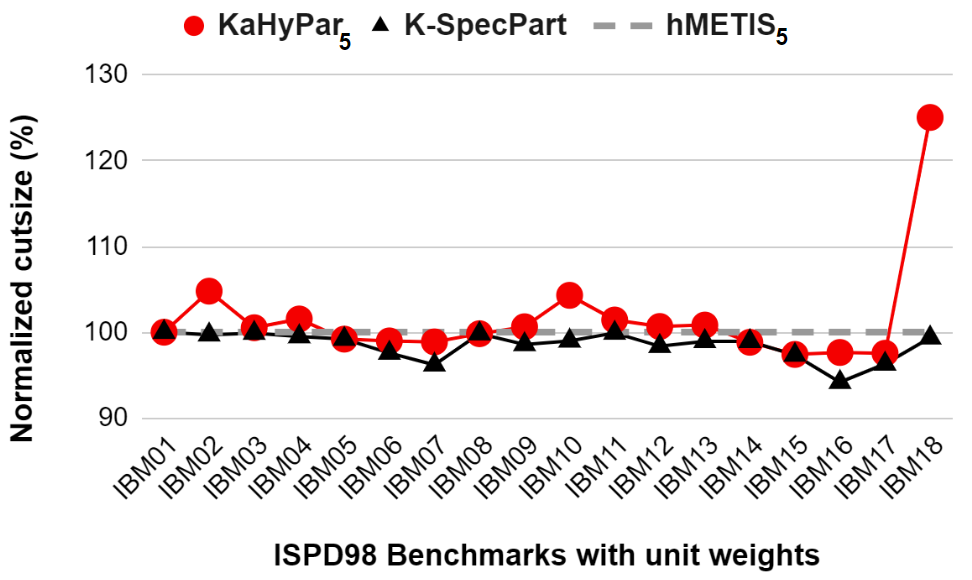}
    \end{subfigure}
    \hfill
    \vspace{5pt}
    \begin{subfigure}[b]{0.48\textwidth}
        \centering
        \includegraphics[width=\textwidth]{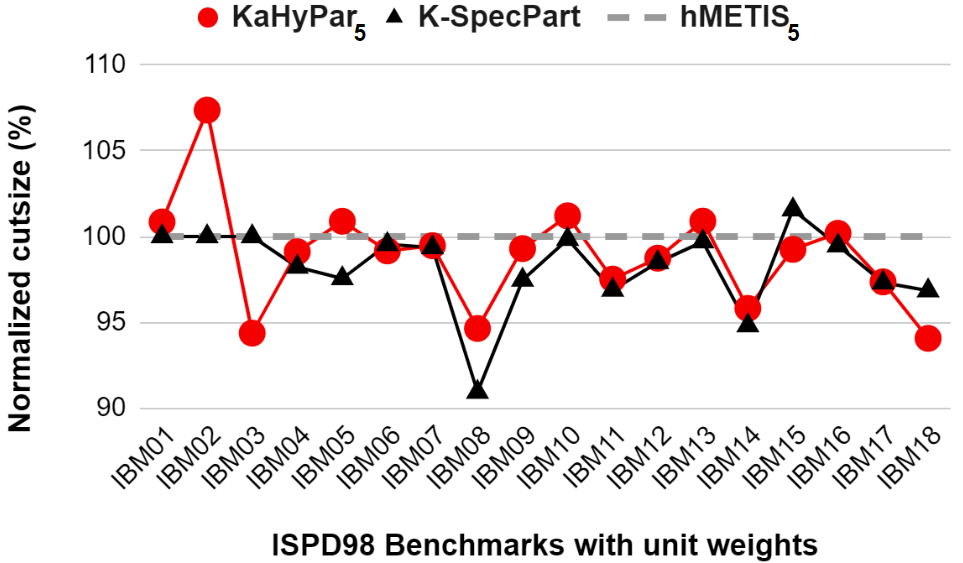}
    \end{subfigure}
    \hfill
    \vspace{5pt}
    \begin{subfigure}[b]{0.48\textwidth}
        \centering
        \includegraphics[width=\textwidth]{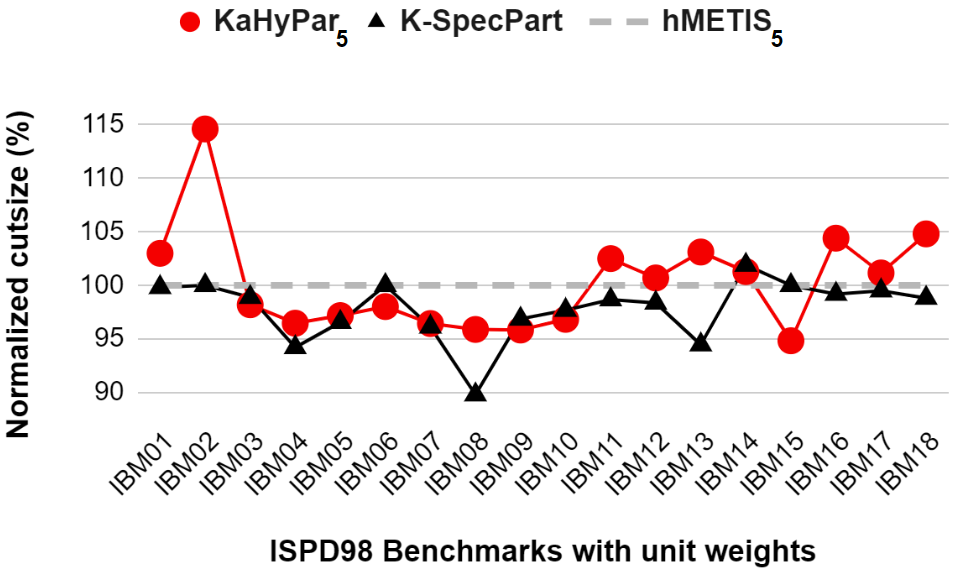}
    \end{subfigure}
    \caption{
    \textcolor{black}{{\em K-SpecPart} results on the {\em ISPD98 Benchmarks}~\cite{Alpert98} with unit vertex
    weights for $\epsilon = 2\%$. Top to bottom: $K=2,3,4$.}}
    \label{fig:IBM_unit_weight}
\end{figure}

\smallskip
\noindent
\textcolor{black}{\textbf{ISPD98 Benchmarks with Unit Weights.} Comparisons with \textit{hMETIS}$_5$ and 
\textit{KaHypar}$_5$ are presented in~Figure \ref{fig:IBM_unit_weight}. 
{\em K-SpecPart} significantly improves over both {\em hMETIS$_{5}$} 
and {\em KaHyPar$_{5}$} on numerous benchmarks for both two-way and  multi-way partitioning. 
Comparisons with {\em hMETIS$_{avg}$} and {\em KaHyPar$_{avg}$} are reported in Table~\ref{table:ispd_unit_wts}.
Each average value is rounded to the nearest tenth (0.1). We observe that {\em K-SpecPart} generates better partitions (\textcolor{black}{$\sim$2\% better on some benchmarks}) than {\em hMETIS$_{avg}$} and {\em KaHyPar$_{avg}$} on the majority of ISPD98 testcases.}

\begin{figure}[!tb]
    \centering
    \begin{subfigure}[b]{0.48\textwidth}
        \centering
        \includegraphics[width=\textwidth]{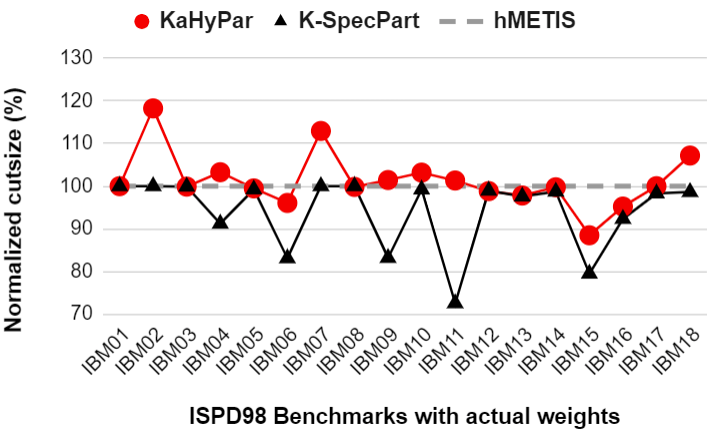}
    \end{subfigure}
    \hfill
    \vspace{5pt}
    \begin{subfigure}[b]{0.48\textwidth}
        \centering
        \includegraphics[width=\textwidth]{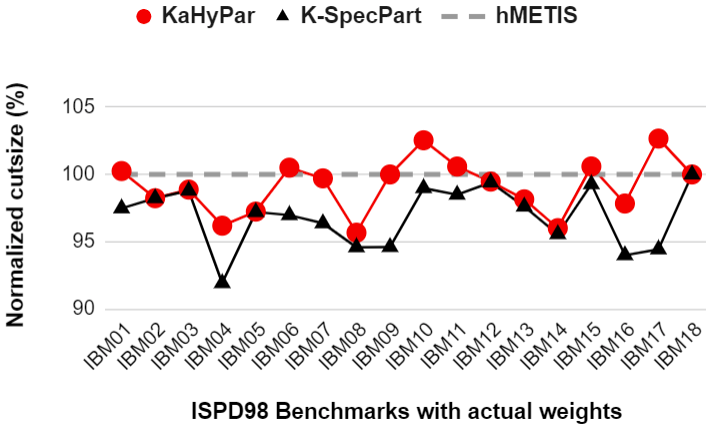}
    \end{subfigure}
    \hfill
    \vspace{5pt}    
    \begin{subfigure}[b]{0.48\textwidth}
        \centering
        \includegraphics[width=\textwidth]{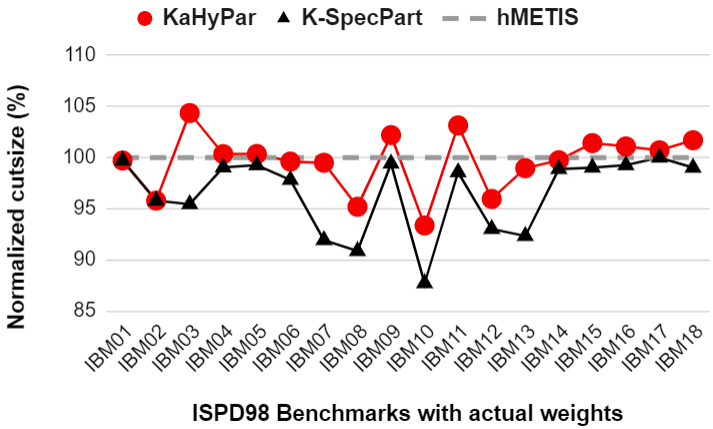}
    \end{subfigure}
    \vspace{5pt}
    \caption{
    \textcolor{black}{{\em K-SpecPart} results on the {\em ISPD98 Benchmarks}~\cite{Alpert98} with actual vertex weights for $\epsilon = 2\%$. Top to bottom: $K=2,3,4$. Cutsizes are normalized with respect to those by \textit{hMETIS}$_5$.}}
    \label{fig:IBM_weight}
\end{figure}

\noindent
\textcolor{black}{\textbf{ISPD98 Benchmarks with Actual Weights.}  The inclusion of weights makes the problem more 
general and potentially more challenging. {Figure \ref{fig:IBM_weight} 
compares {\em K-SpecPart} against {\em hMETIS$_{5}$} and {\em KaHyPar$_{5}$}, while Table~\ref{table:ispd_wts} provides comparisons with {\em hMETIS$_{avg}$} and {\em KaHyPar$_{avg}$}.
We see that {\em K-SpecPart} tends to yield more significant improvements relative to the unit-weight case. For example, for IBM11$_w$, {\em K-SpecPart} generates almost $27$\% improvement 
over {\em hMETIS} and {\em KaHyPar} for $K=2$. We notice similar 
improvements for $K > 2$ as seen on IBM04$_w$ for $K=3$ and IBM10$_w$ for $K=4$.}}

\begin{figure}[!tb]
    \centering
    \begin{subfigure}[b]{0.43\textwidth}
        \centering
        \includegraphics[width=\textwidth]{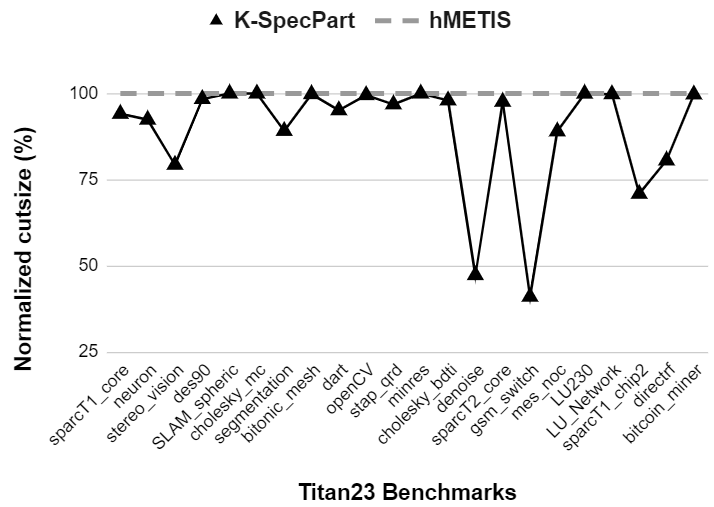}
    \end{subfigure}
    \hfill
    \vspace{5pt}
    \begin{subfigure}[b]{0.43\textwidth}
        \centering
        \includegraphics[width=\textwidth]{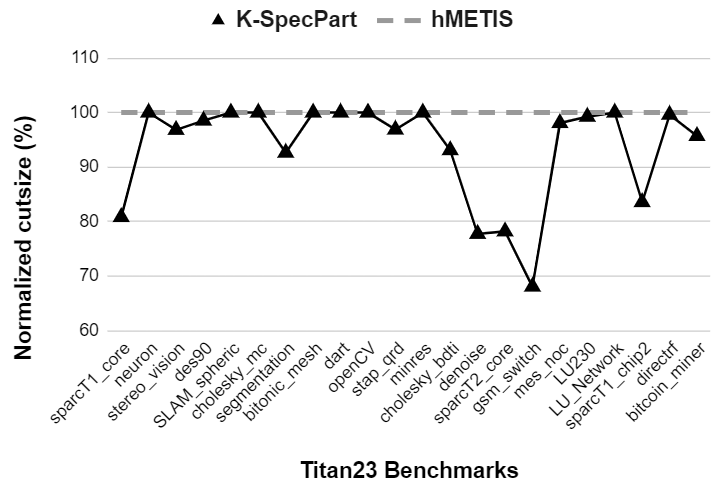}
    \end{subfigure}
    \hfill
    \vspace{5pt}
    \begin{subfigure}[b]{0.43\textwidth}
        \centering
        \includegraphics[width=\textwidth]{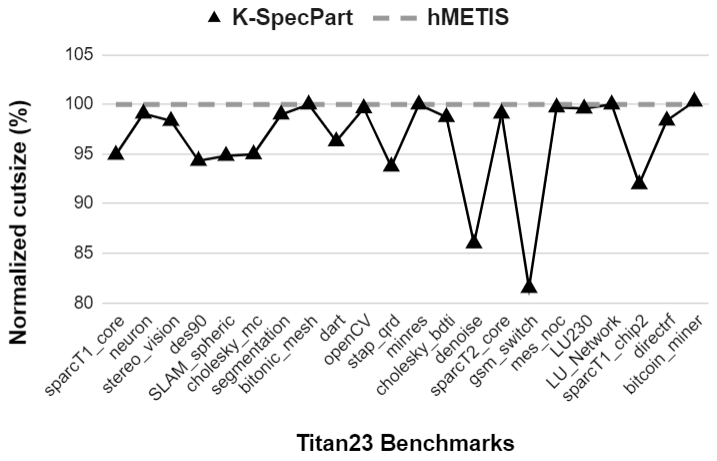}
    \end{subfigure}
    \hfill
        \vspace{5pt}
    \caption{\textcolor{black}{{\em K-SpecPart} results on the {\em Titan23 Benchmarks}~\cite{MurrayWLLB13} 
    for $\epsilon = 2\%$. Top-to-bottom: $K=2,3,4$.  Cutsizes are normalized with respect to those by \textit{hMETIS}$_5$.}}
    \label{fig:titan23}
\end{figure}

\noindent
\textbf{Titan23 Benchmarks.}
The {\em Titan23} benchmarks are interesting not only because they are 
substantially larger than the {\em ISPD98} benchmarks, 
but also because they are generated 
by different, more modern synthesis processes.
In some sense, they provide a ``test of time'' for {\em hMETIS}, as well as 
for {\em KaHyPar} which does not include {\em Titan23} in its experimental 
study~\cite{SebastianTLYCP22}.  \textcolor{black}{Figure~\ref{fig:titan23}~compares \textit{K-SpecPart}  against \textit{hMETIS}$_5$, while Table~\ref{table:titan23} compares with \textit{hMETIS}$_{avg}$.  {Although the {\em K-SpecPart} runtime is still similar to {\em hMETIS}$_5$, the runtime of {\em KaHyPar} on some of these benchmarks is exceedingly long (over two hours), making it unsuitable for any reasonable industrial setting (for more details on runtime, see \cite{TILOS-HGPart}).}
For this reason we do not compare against {\em KaHyPar}.
We observe that {\em K-SpecPart} generates better partitioning solutions 
compared to {\em hMETIS$_{5}$} and {\em hMETIS$_{avg}$}. 
On {\em gsm\_switch} in particular, {\em K-SpecPart} achieves more than 50\% 
better cutsize.}

\subsection{\textcolor{black}{Runtime Remarks}}
\label{sec:runtime}
\textcolor{black}{Our current Julia implementation of \textit{K-SpecPart} has a walltime approximately 5X that of a single \textit{hMETIS} run. \textit{K-SpecPart} does utilize multiple cores, but there is still potential for speedup, in the following ways. (i)~Most of the computational effort is in the embedding generation module. \textcolor{black}{For} $K>2$, \textit{K-SpecPart} employs limited parallelism in the embedding generation module, by solving in parallel the $K$ eigenvector problem instances. The eigensolver has much more potential for parallelism since it relies on sequential and unoptimized sparse matrix-vector multiplications. These can be significantly speeded up on multicore CPUs, GPUs, or other specialized hardware. 
(ii)~In the tree partitioning module, \textit{K-SpecPart} uses parallelism to handle partitioning of multiple trees. The most time-consuming component of this module is the cut distillation algorithm, where there is scope for runtime improvement, especially for larger instances. This can be achieved by implementing the faster LCA algorithm in~\cite{bender2000lca}. 
(iii)~The CPLEX solver can also be accelerated by leveraging the ``warm-start'' feature where a previously computed partitioning solution can be used as an initial solution for the ILP. Furthermore, the CPLEX solver often computes a solution prior to its termination where extra time is spent to produce a computational proof of optimality \cite{CPLEX}. Using a timeout is an option that can accelerate the solver without significantly affecting the quality of the output, but we have not explored this option in \textit{K-SpecPart}.}

\begin{figure}[!t]
    \centering
    \begin{subfigure}[b]{0.43\textwidth}
        \centering
        \includegraphics[width=\textwidth]{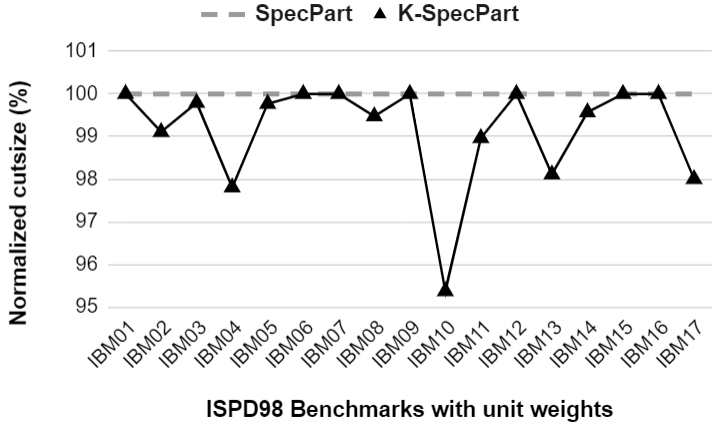}
    \end{subfigure}
    \hfill
    \vspace{5pt}
    \begin{subfigure}[b]{0.43\textwidth}
        \centering
        \includegraphics[width=\textwidth]{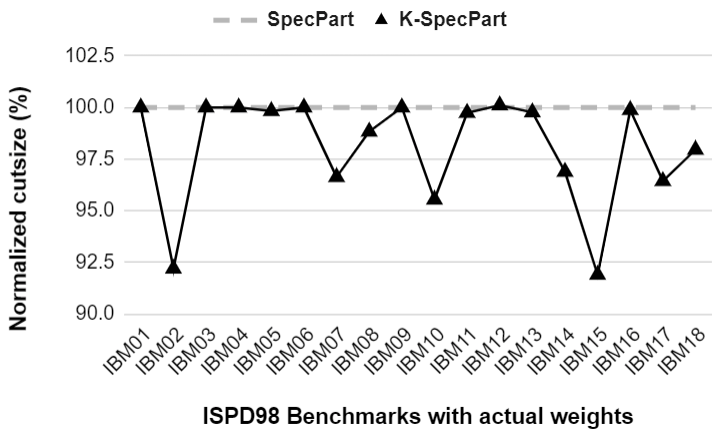}
    \end{subfigure}
    \hfill        
    \vspace{5pt}
    \begin{subfigure}[b]{0.43\textwidth}
        \centering
        \includegraphics[width=\textwidth]{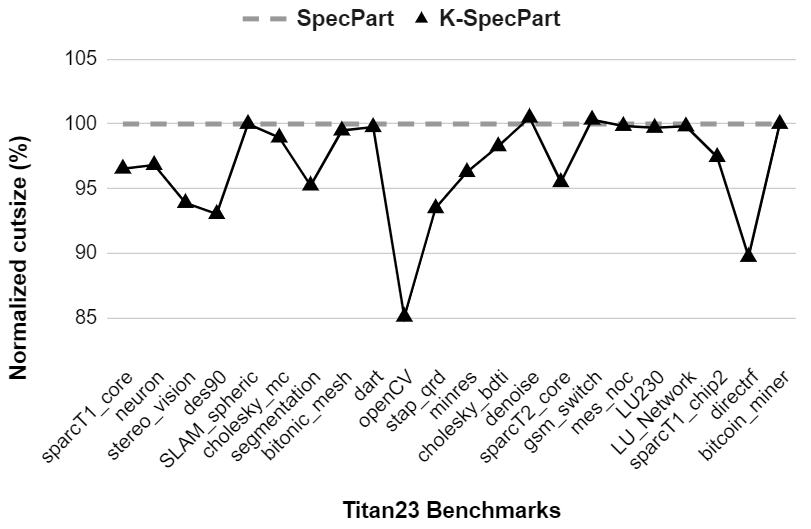}
    \end{subfigure}
    \hfill    
    \vspace{5pt}
    \caption{\textcolor{black}{Comparison of {\em K-SpecPart} and {\em SpecPart} on the bipartitioning problem ($\epsilon$ = $2\%$). Top-to-bottom:  ISPD98 with unit weights, ISPD98 with actual weights and Titan23.}}
    \label{fig:spec_part}
\end{figure}

\subsection{\textcolor{black}{\textit{K-SpecPart} Improvements Over \textit{SpecPart}}}
\label{sec:K-SpecPart_vs_SpecPart}

\textcolor{black}{We have also compared {\em K-SpecPart} for the case $K=2$, against {\em SpecPart}~\cite{SpecPart}.
The results are presented in Figure~\ref{fig:spec_part}.
Although {\em SpecPart} also improves the hint solutions from {\em hMETIS} 
and {\em KaHyPar},
we observe that {\em K-SpecPart} generates significant improvement 
(often in the range of $10$-$15$\%) over {\em SpecPart}
on various benchmarks. This improvement can be attributed to two main 
factors: (i) {\em K-SpecPart} refines the partitioning solutions generated 
from the constructed trees using a FM refinement algorithm; and (ii)  {\em K-SpecPart}
incorporates cut-overlay clustering and ILP-based partitioning in each 
iteration.} 

\subsection{Parameter Validation}
\label{sec:ablation}

\begin{figure}[!t]
    \centering
    {\includegraphics[page=1, width=0.9\linewidth]{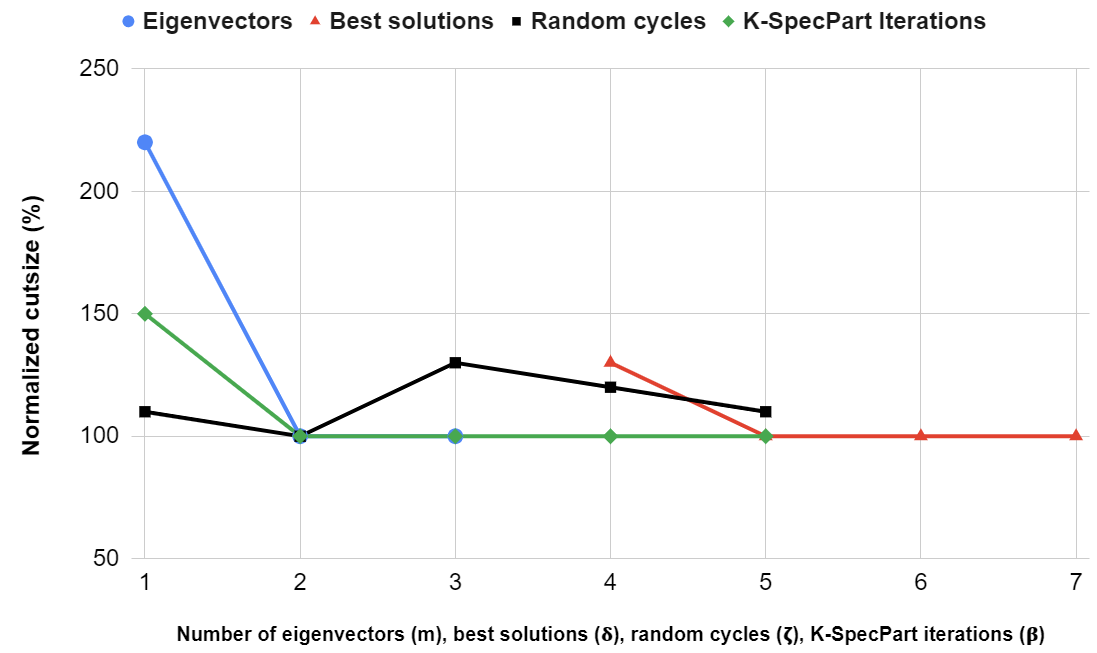}}
\caption{\textcolor{black}{Validation of {\em K-SpecPart} parameters. 
(a) Number of eigenvectors ($m$) sweep;
(b) number of best solutions ($\delta$) sweep;
(c) number of iterations ($\beta$) sweep; and
(d) number of random cycles ($\zeta$) sweep.}}
\label{fig:ablation}
\end{figure}

We now discuss the sensitivity of {\em K-SpecPart} with respect to its parameters, shown in Table~\ref{tab:param}. 
We define the score value as the average improvement of {\em K-SpecPart} 
with respect to {\em hMETIS$_{avg}$}
on benchmarks {\em sparcT1\_core},  {\em  cholesky\_mc}, {\em segmentation}, 
{\em denoise},  {\em gsm\_switch} and {\em  directf}, for $K=2$ and $\epsilon = 5\%$. 
\textcolor{black}{With respect to $\gamma$, we have found that using {\em hMETIS} instead of ILP for partitioning (i.e., setting $\gamma=0$) worsens the score value by $2.68$\%. We have also found that settings of $\gamma>500$ do not improve the score value. For the other parameters, we perform the following experiment. When we vary the value of one parameter (parameter sweep),  the remaining 
parameters are fixed at their default values. 
The results are presented in Figure~\ref{fig:ablation}.} 
From the results of tuning parameters on {\em K-SpecPart} 
we establish that our default parameter setting represents a local minimum in the hyperparameter search space.

\subsection{\textcolor{black}{Effect of Linear Discriminant Analysis (LDA)}}
\label{sec:lda}
\begin{figure}[!h]
    \centering
    {\includegraphics[page=1, width=0.9\linewidth]{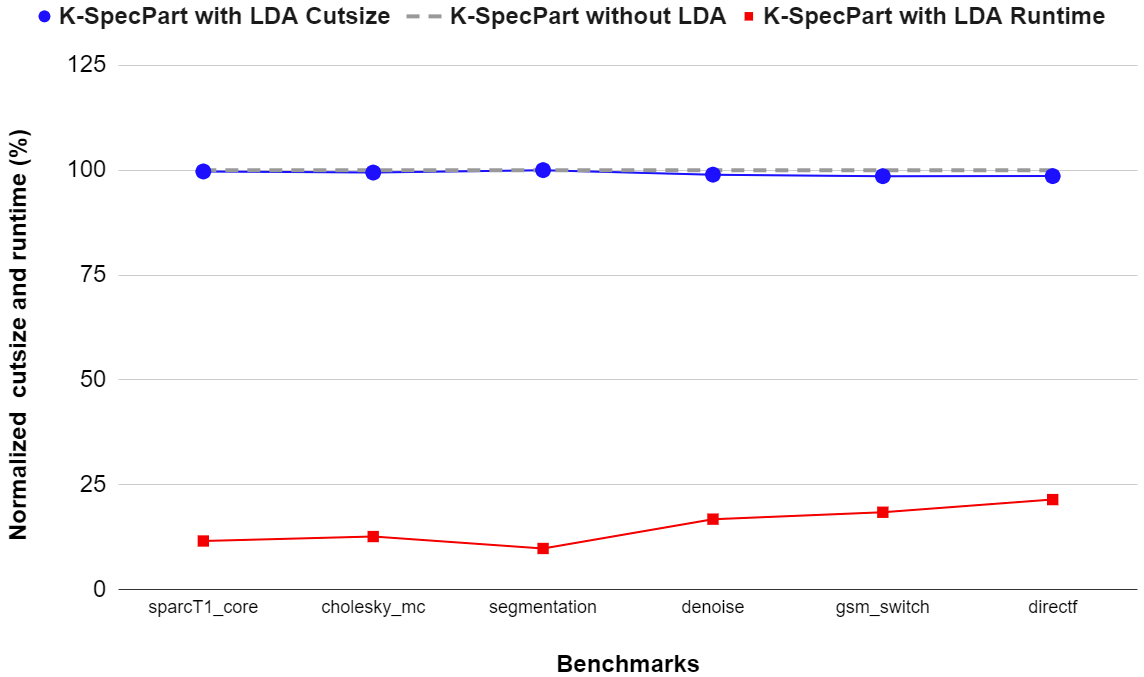}}
\caption{\textcolor{black}{Comparison of cutsize and runtime of {\em K-SpecPart} with LDA and {\em K-SpecPart} without LDA, for $K=4$ and $\epsilon = 2\%$.}}
\label{fig:lda_comparison}
\end{figure}

\textcolor{black}{We have compared the cutsize and runtime of 
{\em K-SpecPart} with LDA, 
and {\em K-SpecPart} without LDA, i.e., utilizing the horizontally stacked eigenvectors $X_{emb}$. 
The result for multi-way partitioning ($K = 4$)
is presented in Figure~\ref{fig:lda_comparison}.
We observe that {\em K-SpecPart} with LDA generates slightly better 
($\sim$1\%) cutsize with significantly faster ($\sim$10X) runtime compared to 
{\em K-SpecPart} without LDA.
However, for the case of bipartitioning ($K=2$) 
we do not observe any significant difference in cutsize when employing LDA.}

\subsection{\textcolor{black}{VILE vs. Recursive Balanced Tree Partitioning}}
\label{sec:vile}

\textcolor{black}{
We additionally compare the ``VILE'' tree partitioning algorithm
(Section \ref{sec:tree_partitioning}) with a balanced tree 
partitioning baseline, based 
on a  recursive two-way cut distilling and partitioning of the tree, 
similar to Section \ref{sec:tree_construction}. During each level of 
recursive partitioning, we dynamically adjust the balance constraint to 
ensure that the final $K$-way partitioning solution satisfies the balance 
constraints (see Section \ref{sec:problem_statement}).
In particular, while executing the $i^{th}$ ($ 1 \leq i \leq K - 1$) level 
bipartitioning, 
the balance constraints associated with the bipartitioning solution 
$S(V_{i0}, V_{i1})$ are:}

\begin{equation}
  \begin{split}
   \textcolor{black}{(\frac{1}{K} - \epsilon)  W  \leq \sum_{v \in V_{i0}}{w_v}  \leq (\frac{1}{K} + \epsilon)W }
  \end{split}
\end{equation}
\begin{equation}
\resizebox{0.4\textwidth}{!}{
      \textcolor{black}{$\sum_{v \in V_{i0}}{w_v} \geq (\sum_{v \in V_{i0}, V_{i1}}{w_v})- (K - i)(\frac{1}{K} + \epsilon)W$}
}
\end{equation}

\begin{figure}
    \centering
    \begin{subfigure}[b]{0.47\textwidth}
    {\includegraphics[page=1, width=0.95\linewidth]{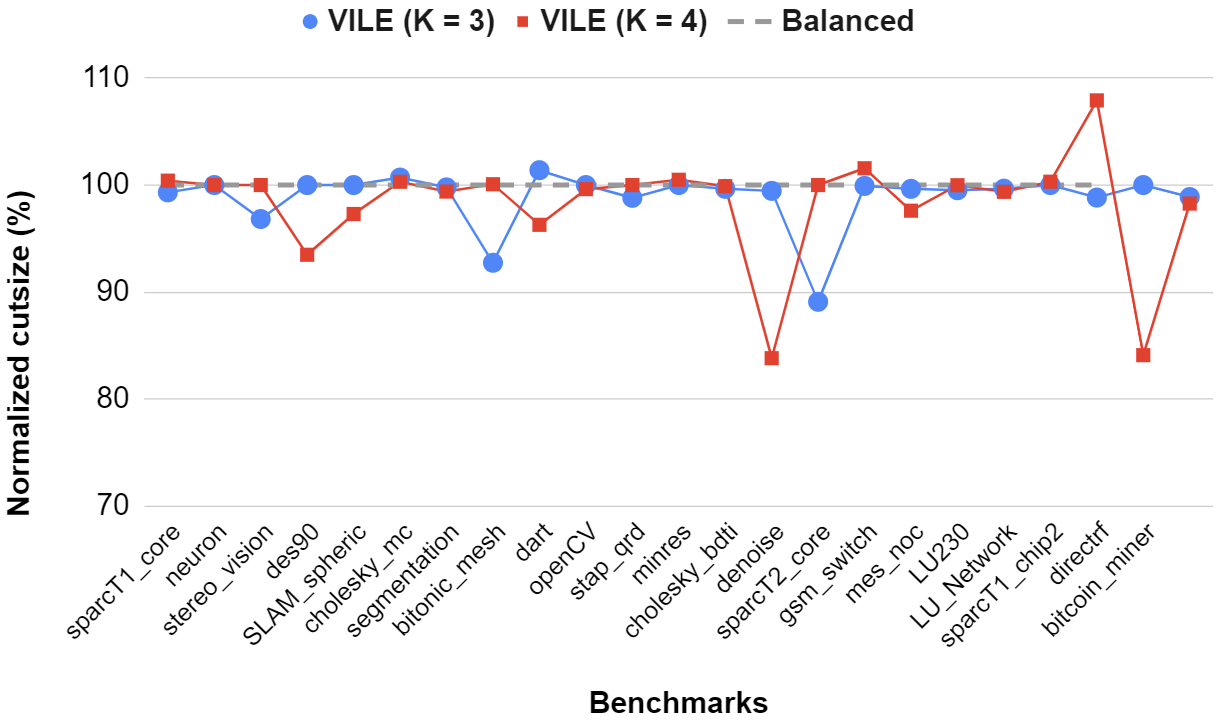}}
    \end{subfigure}
    \hfill
    \vspace{5pt}
\caption{\textcolor{black}{Comparison of ``VILE'' tree partitioning and balanced tree partitioning for $K=3, 4$ and $\epsilon = 2\%$.}}
\label{fig:vile_comparison}
\end{figure}

\textcolor{black}{After obtaining the bipartitioning solution $S(V_{i0}, V_{i1})$, 
we proceed with the $(i + 1)^{th}$ level bipartitioning. 
A comparison of cutsize obtained with ``VILE'' tree partitioning and 
balanced tree partitioning is presented in Figure~\ref{fig:vile_comparison}. The plots are 
normalized with respect to the cutsize obtained with balanced tree 
partitioning. We observe that ``VILE'' tree partitioning yields better 
cutsize (on average $2$\% better) compared to balanced tree partitioning.}

\subsection{\textcolor{black}{Effect of Supervision in \textit{K-SpecPart}}}
\label{sec:supervisioneffect}
\begin{figure}[!tb]
    \centering
    \begin{subfigure}[b]{0.45\textwidth}
        \centering
        \includegraphics[width=\textwidth]{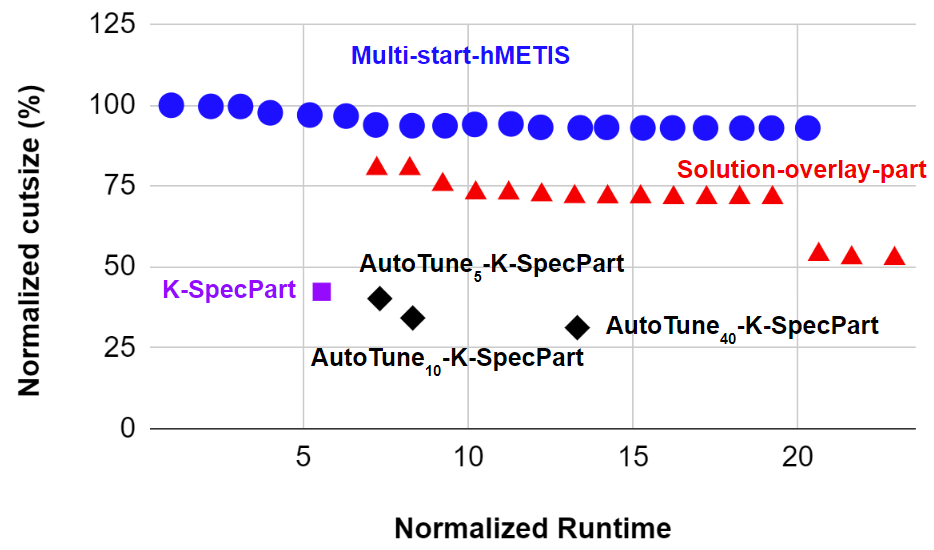}
    \end{subfigure}
    \hfill
    \vspace{5pt}
    \begin{subfigure}[b]{0.45\textwidth}
        \centering
        \includegraphics[width=\textwidth]{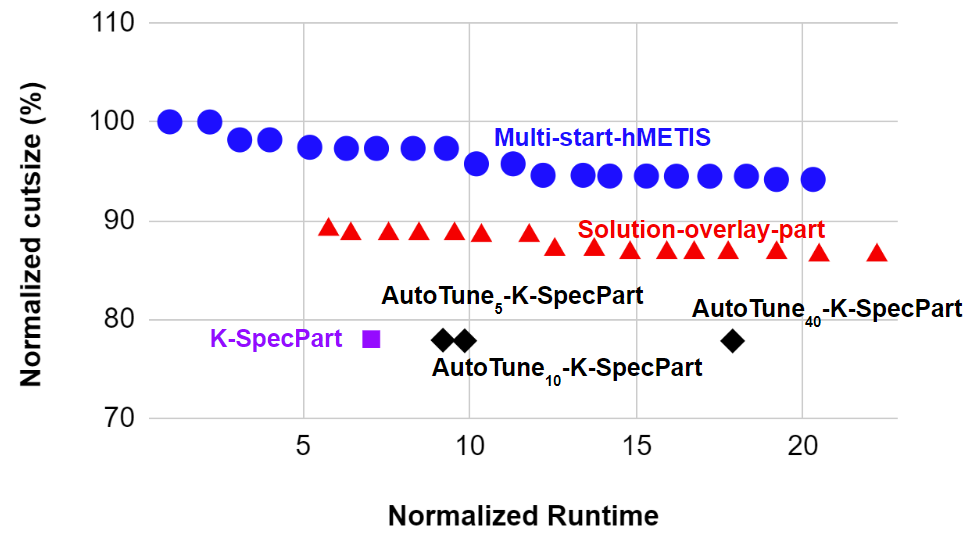}
    \end{subfigure}
    \hfill
    \vspace{5pt}
    \caption{\textcolor{black}{Cutsize versus runtime on  {\em gsm\_switch}, for $\epsilon = 2\%$. \\ Top-to-bottom: $K=2,4$.}}
    \label{fig:qor_runtime}
\end{figure}

In order to show the effect of 
supervision in {\em K-SpecPart},
we run {\em solution ensembling via cut overlay} 
directly on candidate solutions,
which are generated by running {\em hMETIS} multiple times with different 
random seeds.
The flow is as follows.
(i)~We generate candidate solutions $\{S_1, S_2, ..., S_{\psi}\}$ by running 
{\em hMETIS} {\em $\psi$} times with different random seeds,
and report the best cutsize {\em Multi-start-hMETIS}. Here $\psi$ is an 
integer parameter ranging from 1 to 20.
(ii)~We run {\em solution ensembling via cut overlay} 
directly on the best five solutions from $\{S_1, S_2, ..., S_{\psi}\}$
and report the cutsize {\em Solution-overlay-part}.
For each value of $\psi$, we run this flow $100$ times and report the 
average result in 
Figure \ref{fig:qor_runtime}.
\textcolor{black}{We observe that {\em Solution-overlay-part} is much better than 
{\em Multi-start-hMETIS},
and that {\em K-SpecPart} generates superior solutions in 
less runtime compared to {\em Multi-start-hMETIS} and 
{\em Solution-overlay-part}.
This suggests that supervision is an important component 
of {\em K-SpecPart}.}

\noindent
\subsection{Solution Enhancement by Autotuning.}
\label{sec:autotuning}
{\em hMETIS} has parameters whose settings may significantly
impact the quality of generated partitioning solutions.
We use Ray~\cite{Ray} to tune the following parameters of {\em hMETIS}:
CType with possible values $\{ 1, 2, 3, 4, 5 \}$,
RType with possible values $\{ 1, 2, 3 \}$,
Vcycle with possible values $\{ 1, 2, 3 \}$,
and Reconst with possible values $\{0, 1 \}$.
The search algorithm we use in Ray~\cite{Ray} is {\em HyperOptSearch}.
We set the number of trials, i.e., total number of runs of {\em hMETIS}
launched by Ray, to 5, 10 and 40.
We set the number of threads to 10 to reduce the runtime (elapsed walltime).  
Here we normalize the cutsize and runtime to that of running {\em hMETIS} 
once with default random seed. \textcolor{black}{Autotuning increases the 
runtime for {\em hMETIS} and computes a better hint $S_{init}$; \textcolor{black}{it leads to} 
a further $2\%$ cutsize improvement from {\em K-SpecPart} on 
{\em gsm\_switch} for $K=2$ and $K=4$.}

\section{Conclusion and Future Directions}
\label{sec:conclusion}
\textcolor{black}{We have proposed {\em K-SpecPart}, the first general supervised framework
for hypergraph multi-way partitioning solution improvement. 
Our experimental results demonstrate the superior performance of 
{\em K-SpecPart} in comparison to traditional multilevel partitioners, 
while maintaining comparable runtimes for both bipartitioning and multi-way 
partitioning. The findings from {\em SpecPart} and {\em K-SpecPart} indicate 
that the partitioning problem may not be as comprehensively solved as 
previously believed, and that substantial advancements may yet remain to be discovered.}
\textcolor{black}{{\em K-SpecPart} can be integrated 
with the internal levels of multilevel
partitioners; producing improved solutions on each level may lead
to further improved solutions.} Furthermore, we believe that 
the {\em cut-overlay clustering}
and LDA-based embedding generation hold independent interest and are
amenable to machine learning techniques.

\smallskip
\noindent {\textbf{Acknowledgments.}} \textcolor{black}{We thank Dr. Grigor Gasparyan for 
sharing his thoughts on {\em K-SpecPart}.} This work was partially supported 
by NSF grants CCF-2112665, CCF-2039863 and CCF-1813374 and 
by DARPA HR0011-18-2-0032.

\balance

\clearpage

\begin{table*}
\centering
  \resizebox{1.5\columnwidth}{!}{
  \begin{tabular}{|c|c|c|c|c|c|c|c|c|c|c|c|}
    \hline
    \multirow{2}{*}{} &
      \multicolumn{2}{c}{\textbf{Statistics}} &
      \multicolumn{3}{c}{\textbf{$K = 2$}} &
      \multicolumn{3}{c}{\textbf{$K = 3$}} &
      \multicolumn{3}{c|}{\textbf{$K = 4$}}\\
      \hline
    \textbf{Benchmark} & \textbf{ $|V|$} & \textbf{ $|E|$} & \textbf{{\em hM$_{avg}$}} & \textbf{{\em KHPr$_{avg}$}} &\textbf{{\em K-SP}} & \textbf{{\em hM$_{avg}$}} & \textbf{{\em KHPr$_{avg}$}} & \textbf{{\em K-SP}} & \textbf{{\em hM$_{avg}$}} & \textbf{{\em KHPr$_{avg}$}} & \textbf{{\em K-SP}} \\
    \hline  
    IBM01 & $12752$ & $14111$ & $203.0$ & $203.0$ & $203$ & $352.0$ & $355.2$ & $352$ & $503.7$ & \textcolor{blue}{$493.3$} & $522$ \\
    \hline
    IBM02 & $19601$ & $19584$ & \textcolor{blue}{$331.5$} & $350$ & $333$ & $339.4$ & $357.2$ & \textcolor{blue}{$339$} & \textcolor{blue}{$676.5$} & $701.3$ & $706$ \\
    \hline
    IBM03 & $23136$ & $27401$ & $958.3$ & $957.2$ & \textcolor{blue}{$957$} & $1544.2$ & \textcolor{blue}{$1482.1$} & $1480$ & $1701.6$ & $1693.9$ & \textcolor{blue}{$1690$} \\
    \hline
    IBM04 & $27507$ & $31970$ & $581.3$ & $581.3$ & \textcolor{blue}{$580$} & \textcolor{blue}{$1199.6$} & $1203.6$ & $1212$ & $1669.4$ & $1631.2$ & \textcolor{blue}{$1626$} \\
    \hline
    IBM05 & $29347$ & $28446$ & $1728.6$ & $1718.9$ & \textcolor{blue}{$1716$} & $2645.2$ & $2642.4$ & \textcolor{blue}{$2635$} & $3031.2$ & $2967.8$ & \textcolor{blue}{$2946$} \\
    \hline
    IBM06 & $32498$ & $34826$ & \textcolor{blue}{$974.3$} & $977.2$ & $976$ & $1306.7$ & \textcolor{blue}{$1298.2$} & $1305$ & $1517.9$ & $1515.2$ &  \textcolor{blue}{$1476$} \\
    \hline
    IBM07 & $45926$ & $48117$ & \textcolor{blue}{$910.3$} & $913.2$ & $935$ & $1882.4$ & $1873.4$ & \textcolor{blue}{$1846$} & $2201.4$ & $2165.2$ & \textcolor{blue}{$2154$} \\
    \hline
    IBM08 & $51309$ & $50513$ & $1141.2$ & $1140.2$ & \textcolor{blue}{$1140$} & $2056.2$ & \textcolor{blue}{$2011.9$} & $2037$ & $2401.5$ & $2348.7$ & \textcolor{blue}{$2328$} \\
    \hline
    IBM09 & $53395$ & $60902$ & $625.3$ & $625.9$ & \textcolor{blue}{$620$} & $1404.1$ & $1407.2$ & \textcolor{blue}{$1384$} & $1734.2$ & \textcolor{blue}{$1675.1$} & $1676$ \\
    \hline
    IBM10 & $69429$ & $75196$ & $1280.3$ & $1327.7$ & \textcolor{blue}{$1257$} & $1911.8$ & $1904.3$ & \textcolor{blue}{$1880$} & $2445.2$ & \textcolor{blue}{$2372.9$} & $2400$ \\
    \hline
    IBM11 & $70558$ & $81454$ & $1052.6$ & $1066.5$ & \textcolor{blue}{$1051$} & $1808.5$ & \textcolor{blue}{$1789.6$} & $1843$ & $2458.9$ & $2465.8$ & \textcolor{blue}{$2452$} \\
    \hline
    IBM12 & $71076$ & $77240$ & $1947.6$ & $1961.9$ & \textcolor{blue}{$1937$} & $2817.3$ & $2816.3$ & \textcolor{blue}{$2791$} & $3870.4$ & $3894.2$ & \textcolor{blue}{$3844$} \\
    \hline
    IBM13 & $84199$ & $99666$ & $844.3$ & $848.4$ & \textcolor{blue}{$832$} & $1347.8$ & $1345.3$ & \textcolor{blue}{$1335$} & $1913.2$ & $1941.7$ & \textcolor{blue}{$1904$} \\
    \hline
    IBM14 & $147605$ & $152772$ & $1875.1$ & \textcolor{blue}{$1849.6$} & $1850$ & $2789.9$ & \textcolor{blue}{$2607.5$} & $2710$ & \textcolor{blue}{$3401.4$} & $3453.7$ & $3475$ \\
    \hline
    IBM15 & $161570$ & $186608$ & $2817.2$ & $2741.7$ & \textcolor{blue}{$2741$} & $4200.8$ & \textcolor{blue}{$4114.2$} & $4333$ & $4870.7$ & \textcolor{blue}{$4627.5$} & $4720$ \\
    \hline
    IBM16 & $183484$ & $190048$ & $1925.6$ & $2017.6$ & \textcolor{blue}{$1921$} & $3169.3$ & $3107.3$ & \textcolor{blue}{$3062$} & \textcolor{blue}{$4045.2$} & $4216.7$ & $4060$ \\
    \hline
    IBM17 & $185495$ & $189581$ & $2364.5$ & $2332.2$ & \textcolor{blue}{$2307$} & $4550.1$ & $4305.1$ & \textcolor{blue}{$4248$} & $5634.6$ & $5738.9$ & \textcolor{blue}{$5583$} \\
    \hline
    IBM18 & $210613$ & $201920$ & $1531.6$ & $1893.4$ & \textcolor{blue}{$1523$} & $2543.9$ & $2487.6$ & \textcolor{blue}{$2401$} & $2949.4$ & $2984.5$ & \textcolor{blue}{$2918$} \\
    \hline 
    \end{tabular}
  }
  \caption{\textcolor{black}{Comparison of {\em hMETIS}, {\em KaHyPar} and {\em K-SpecPart} on ISPD98 benchmarks with unit vertex weights for multi-way partitioning with number of blocks ($K$) = $2, 3, 4$ and imbalance factor ($\epsilon$) = $2\%$}. }
\label{table:ispd_unit_wts}
\end{table*}

\begin{table*}
\centering
 \resizebox{1.5\columnwidth}{!}{
  \begin{tabular}{|c|c|c|c|c|c|c|c|c|c|c|c|}
    \hline
    \multirow{2}{*}{} &
      \multicolumn{2}{c}{\textbf{Statistics}} &
      \multicolumn{3}{c}{\textbf{$K = 2$}} &
      \multicolumn{3}{c}{\textbf{$K = 3$}} &
      \multicolumn{3}{c|}{\textbf{$K = 4$}}\\
      \hline
    \textbf{Benchmark} & \textbf{ $|V|$} & \textbf{ $|E|$} & \textbf{{\em hM$_{avg}$}} & \textbf{{\em KHPr$_{avg}$}} &\textbf{{\em K-SP}} & \textbf{{\em hM$_{avg}$}} & \textbf{{\em KHPr$_{avg}$}} & \textbf{{\em K-SP}} & \textbf{{\em hM$_{avg}$}} & \textbf{{\em KHPr$_{avg}$}} & \textbf{{\em K-SP}} \\
    \hline  
    $IBM01_{w}$ & $12752$ & $14111$ & $215.0$ & $215.1$ & $215$ & $389.1$ & $396.4$ & 
    \textcolor{blue}{$387$} & $349.0$ & \textcolor{blue}{$347.6$} & $349$ \\
    \hline
    $IBM02_{w}$ & $19601$ & $19584$ & $324.6$ & $349.4$ & \textcolor{blue}{$296$} & $340.9$ & $354.3$ & \textcolor{blue}{$334$} & $548.4$ & $565.7$ & \textcolor{blue}{$524$} \\
    \hline
    $IBM03_{w}$ & $23136$ & $27401$ & $958.2$ & $957.2$ & \textcolor{blue}{$957$} & $1249.8$ & $1471.7$ & \textcolor{blue}{$1230$} & $1496.2$ & $1473.1$ & \textcolor{blue}{$1447$} \\
    \hline
    $IBM04_{w}$ & $27507$ & $31970$ & $584.7$ & $582.4$ & \textcolor{blue}{$529$} & $899.6$ & $805.4$ & \textcolor{blue}{$797$} & $1572.4$ & \textcolor{blue}{$1532.4$} & $1446$ \\
    \hline
    $IBM05_{w}$ & $29347$ & $28446$ & $1728.9$ & \textcolor{blue}{$1716.9$} & $1721$ & $2640.1$ & \textcolor{blue}{$2641.2$} & $2642$ & $3035.7$ & \textcolor{blue}{$2960.5$} & $3061$ \\
    \hline
    $IBM06_{w}$ & $32498$ & $34826$ & $969.3$ & $979.7$ & \textcolor{blue}{$845$} & $997.2$ & $998.1$ & \textcolor{blue}{$963$} & $1263.0$ & $1314.8$ & \textcolor{blue}{$1261$} \\
    \hline
    $IBM07_{w}$ & $45926$ & $48117$ & $812.3$ & $824.5$ & \textcolor{blue}{$803$} & $1375.2$ & $1367.1$ & \textcolor{blue}{$1328$} & $1902.4$ & $1897.6$ & \textcolor{blue}{$1824$} \\
    \hline
    $IBM08_{w}$ & $51309$ & $50513$ & $1142.4$ & \textcolor{blue}{$1140.2$} & $1182$ & $1844.3$ & $1794.2$ & \textcolor{blue}{$1749$} & $2407.9$ & $2350.9$ & \textcolor{blue}{$2268$} \\
    \hline
    $IBM09_{w}$ & $53395$ & $60902$ & $626.7$ & $625.9$ & \textcolor{blue}{$519$} & $1407.2$ & $1404.1$ & \textcolor{blue}{$1334$} & \textcolor{blue}{$1530.3$} & $1675.9$ & $1535$ \\
    \hline
    $IBM10_{w}$ & $69429$ & $75196$ & $1079.8$ & $1329.1$ & \textcolor{blue}{$1028$} & $1576.0$ & $1655.5$ & \textcolor{blue}{$1560$} & $2447.4$ & $2328.4$ & \textcolor{blue}{$2143$} \\
    \hline
    $IBM11_{w}$ & $70558$ & $81454$ & $845.2$ & $863.9$ & \textcolor{blue}{$763$} & $1509.2$ & \textcolor{blue}{$1501.3$} & $1508$ & $2069.5$ & $2078.9$ & \textcolor{blue}{$2068$} \\
    \hline
    $IBM12_{w}$ & $71076$ & $77240$ & \textcolor{blue}{$1947.3$} & $1962.9$ & $1967$ &  \textcolor{blue}{$2814.4$} & $2816.5$ & $3185$ & $3859.7$ & $3793.4$ & \textcolor{blue}{$3613$} \\
    \hline
    $IBM13_{w}$ & $84199$ & $99666$ & $847.8$ & $848.8$ & \textcolor{blue}{$846$} & $1639.1$ & \textcolor{blue}{$1624.4$} & $1636$ & $1795.1$ & $1840.2$ & \textcolor{blue}{$1784$} \\
    \hline
    $IBM14_{w}$ & $147605$ & $152772$ & $1872.7$ & \textcolor{blue}{$1855.2$} & $1929$ & $2814.5$ & \textcolor{blue}{$2604.1$} & $2780$ & $3374.3$ & \textcolor{blue}{$3251.2$} & $3455$ \\
    \hline
    $IBM15_{w}$ & $161570$ & $186608$ & $2797.1$ & $2741.3$ & \textcolor{blue}{$2474$} & $3864.5$ & $3915.6$ & \textcolor{blue}{$3836$} & $4805.6$ & \textcolor{blue}{$4633.3$} & $4758$ \\
    \hline
    $IBM16_{w}$ & $183484$ & $190048$ & $1656.4$ & $1689.2$ & \textcolor{blue}{$1660$} & $2942.3$ & $2981.9$ & \textcolor{blue}{$2742$} & \textcolor{blue}{$3676.5$} & $3892.1$ & $3729$ \\
    \hline
    $IBM17_{w}$ & $185495$ & $189581$ & $2369.5$ & $2334.5$ & \textcolor{blue}{$2301$} & $3589.3$ & $3671.2$ & \textcolor{blue}{$3580$} & \textcolor{blue}{$5630.2$} & $5729.1$ & $5738$ \\
    \hline
    $IBM18_{w}$ & $210613$ & $201920$ & \textcolor{blue}{$1572.1$} & $1930.3$ & $1579$ & $2503.3$ & \textcolor{blue}{$2482.1$} & $2836$ & \textcolor{blue}{$2947.9$} & $2979.6$ & $3209$ \\
    \hline 
    \end{tabular}
}
  \caption{\textcolor{black}{Comparison of {\em hMETIS}, {\em KaHyPar} and {\em K-SpecPart} on ISPD98 benchmarks with actual weights for multi-way partitioning with number of blocks ($K$) = $2, 3, 4$ and imbalance factor ($\epsilon$) = $2\%$.}}
    \label{table:ispd_wts}
\end{table*}

\begin{table*}
\centering
  \begin{tabular}{|c|c|c|c|c|c|c|c|c|}
    \hline
    \multirow{2}{*}{} &
      \multicolumn{2}{c}{\textbf{Statistics}} &
      \multicolumn{2}{c}{\textbf{$K = 2$}} &
      \multicolumn{2}{c}{\textbf{$K = 3$}} &
      \multicolumn{2}{c|}{\textbf{$K = 4$}}\\
      \hline
    \textbf{Benchmark} & \textbf{ $|V|$} & \textbf{ $|E|$} & \textbf{{\em hM$_{avg}$}} & \textbf{{\em K-SP}} & \textbf{{\em hM$_{avg}$}} & \textbf{{\em K-SP}} & \textbf{{\em hM$_{avg}$}} & \textbf{{\em K-SP}} \\
    \hline  
    sparcT1\_core & $91976$ & $92827$ & $982.2$ & \textcolor{blue}{$977$} & $2187.9$ & \textcolor{blue}{$1889$} & $2532.3$ & \textcolor{blue}{$2492$}\\
    \hline
    neuron & $92290$ & $125305$ & $245.0$ & \textcolor{blue}{$244$} & \textcolor{blue}{$371.6$} & $396$ & $431.5$ & \textcolor{blue}{$431$} \\
    \hline
    stereo\_vision & $94050$ & $127085$ & $171.0$ & \textcolor{blue}{$169$} & \textcolor{blue}{$332.7$} & $336$ & \textcolor{blue}{$440.2$} & $475$ \\
    \hline
    des90 & $111221$ & $139557$ & $376.8$ & \textcolor{blue}{$374$} &$536.5$ &  \textcolor{blue}{$535$} & \textcolor{blue}{$695.5$} & $747$  \\
    \hline
    SLAM\_spheric & $113115$ & $142408$ & $1061.0$ & $1061$ & $2797.1$ & \textcolor{blue}{$2720$} & $3371.1$ & \textcolor{blue}{$3241$} \\
    \hline
    cholesky\_mc & $113250$ & $144948$ & $282.0$ & \textcolor{blue}{$282$} & $886.5$ & \textcolor{blue}{$864$} & \textcolor{blue}{$982.8$} & $984$ \\
    \hline
    segmentation & $138295$ & $179051$ & $120.1$ & \textcolor{blue}{$120$} & $476.1$ &  \textcolor{blue}{$453$} & $496.3$ & \textcolor{blue}{$490$} \\
    \hline
    bitonic\_mesh & $192064$ & $235328$ & $585.2$ & \textcolor{blue}{$584$} & $895.0$ & $895$ & \textcolor{blue}{$1304.4$} & $1311$ \\
    \hline
    dart & $202354$ & $223301$ & $837.0$ & \textcolor{blue}{$805$} & \textcolor{blue}{$1189.9$} & $1243$ & $1429.9$ & \textcolor{blue}{$1401$} \\
    \hline
    openCV & $217453$ & $284108$ & $435.4$ & \textcolor{blue}{$434$} &  \textcolor{blue}{$501.8$} & $525$ & $526.2$ & \textcolor{blue}{$522$}  \\
    \hline
    stap\_qrd & $240240$ & $290123$ & \textcolor{blue}{$377.4$} & $464$ & $501.2$ & \textcolor{blue}{$497$} & $714.5$ & \textcolor{blue}{$674$} \\
    \hline
    minres & $261359$ & $320540$ & $207.0$ & $207$ & $309.0$ & $309$ & $407.0$ & $407$ \\
    \hline
    cholesky\_bdti & $266422$ & $342688$ & $1156.0$ & \textcolor{blue}{$1136$} & $1769.2$ & \textcolor{blue}{$1755$} & $1874.4$ & \textcolor{blue}{$1865$} \\
    \hline
    denoise & $275638$ & $356848$ & $496.9$ & \textcolor{blue}{$418$} & $952.8$ & \textcolor{blue}{$915$} & $1172.1$ & \textcolor{blue}{$1001$} \\
    \hline
    sparcT2\_core & $300109$ & $302663$ & $1220.7$ & \textcolor{blue}{$1188$} & $2827.2$ & \textcolor{blue}{$2249$} & \textcolor{blue}{$3323.5$} & $3558$ \\
    \hline
    gsm\_switch & $493260$ & $507821$ & $4235.3$ & \textcolor{blue}{$1833$} & $4148.6$ & \textcolor{blue}{$3694$} & $5169.2$ & \textcolor{blue}{$4404$} \\
    \hline
    mes\_noc & $547544$ & $577664$ & $634.6$ & \textcolor{blue}{$633$} & $1164.3$ & \textcolor{blue}{$1125$} & \textcolor{blue}{$1314.7$} & $1346$ \\
    \hline
    LU230 & $574372$ & $669477$ & \textcolor{blue}{$3333.3$} & $3363$ & $4549.5$ & \textcolor{blue}{$4548$} & $6325.3$ & \textcolor{blue}{$6310$} \\
    \hline
    LU\_Network & $635456$ & $726999$ & $524.0$ & $524$ & \textcolor{blue}{$787.1$} & $882$ & $1495.6$ & \textcolor{blue}{$1417$}  \\
    \hline
    sparcT1\_chip2 & $820886$ & $821274$ & $914.2$ & \textcolor{blue}{$876$} & $1453.4$ & \textcolor{blue}{$1404$} & $1609.8$ & \textcolor{blue}{$1601$} \\
    \hline
    directrf & $931275$ & $1374742$ & $602.6$ & \textcolor{blue}{$515$} & \textcolor{blue}{$728.2$} & $762$ & $1103.6$ & \textcolor{blue}{$1092$} \\
    \hline
    bitcoin\_miner & $1089284$ & $1448151$ & \textcolor{blue}{$1514.1$} & $1562$ & $1944.8$ & \textcolor{blue}{$1917$}
    & \textcolor{blue}{$2605.4$} & $2737$ \\
    \hline  
    \end{tabular}
  \caption{\textcolor{black}{Comparison of {\em hMETIS} and {\em K-SpecPart} on Titan23 benchmarks for multi-way partitioning with number of blocks ($K$) = $2, 3, 4$ and imbalance factor ($\epsilon$) = $2\%$.}}
\label{table:titan23}
\end{table*}


\end{document}